%% file: main.tex
\pdfoutput=1
\documentclass[conference]{IEEEtran}
\usepackage{times}

\usepackage[numbers,sort&compress]{natbib}
\usepackage{multicol}
\input{math_commands}

\input{packages}
\input{macro}

\title{\method: Legged Robot Vision-Language-Action Model for Navigation}
\author{
An-Chieh Cheng$^{1,\ast}$ \quad\quad Yandong Ji$^{1,\ast}$ \quad\quad Zhaojing Yang$^{2,\ast}$ \quad\quad Zaitian Gongye$^{1}$ \quad\quad Xueyan Zou$^{1}$ \quad\quad \\ Jan Kautz$^{3}$ \quad\quad Erdem Bıyık$^{2}$ \quad\quad Hongxu Yin$^{3,\dagger}$ \quad\quad Sifei Liu$^{3,\dagger}$ \quad\quad Xiaolong Wang$^{1,3,\dagger}$ \\ 
\normalsize{ $^1$UC San Diego \quad $^2$USC \quad $^3$NVIDIA } \\
\small{\href{https://navila-bot.github.io}{\texttt{https://navila-bot.github.io}} }}

\begin{document}

\twocolumn[{%
\renewcommand\twocolumn[1][]{#1}%

\maketitle

\begin{center}
    \centering 
    \vspace{-2.7em}
    \includegraphics[width=\linewidth]{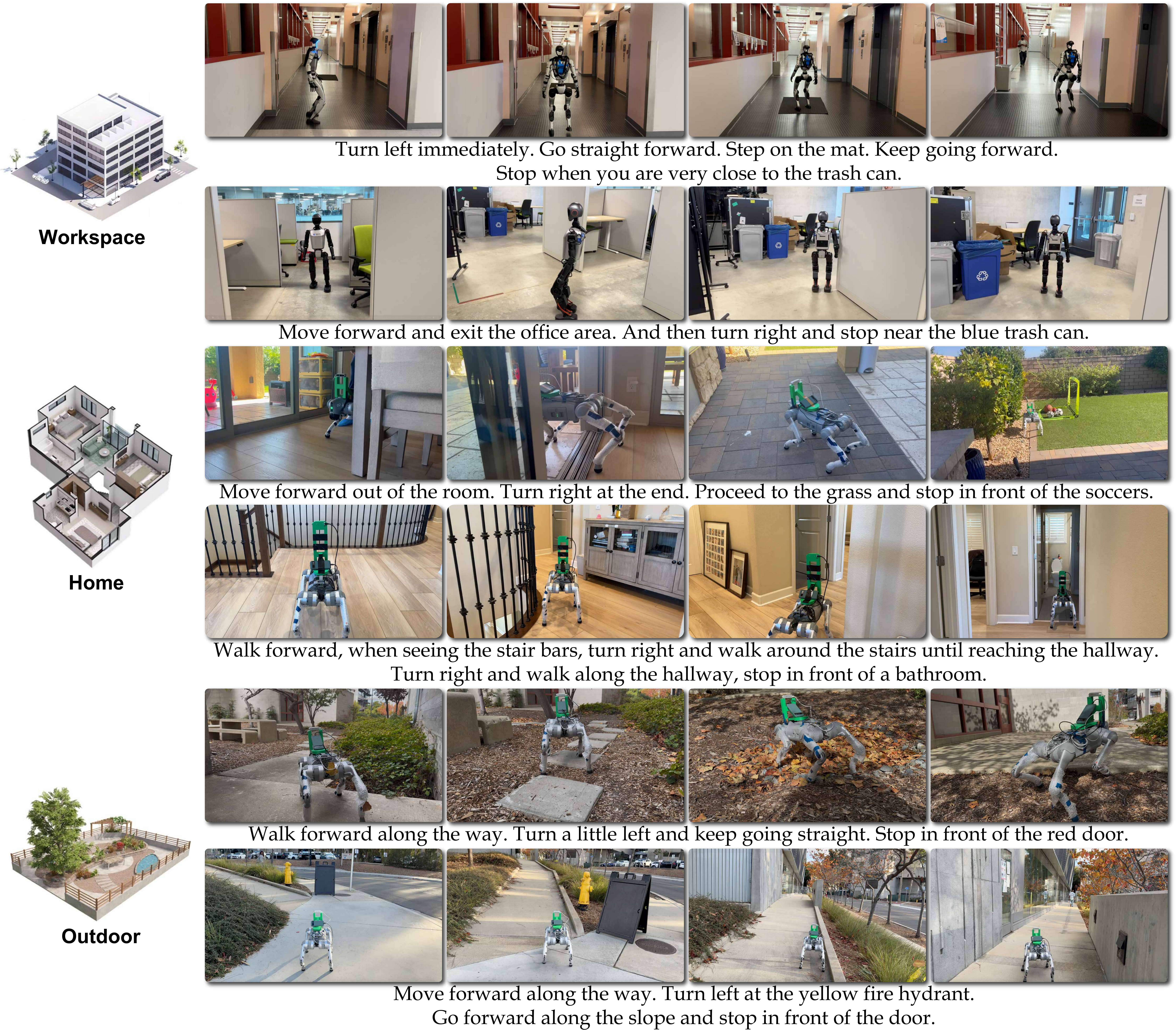}
    \vspace{-1.5em}
    \captionof{figure}{Real-world demonstration of NaVILA: Upon receiving human instructions, NaVILA uses a vision-language model to process RGB video frames and employs locomotion skills to execute the task on a robot. The robot successfully handles long-horizon navigation tasks and operates safely in challenging environments.}
    \label{fig:teaser}
    \vspace{-1.2em}
\end{center}
  }]

\blfootnote{$*$ Equal contribution, ordered alphabetically. $\dag$ Equal advising. }

\IEEEpeerreviewmaketitle

\input{sections/00-Abstract}

\input{sections/01-Introduction}
\input{sections/03-Method2}

\input{sections/04-Experiments}
\input{sections/02-Related-Work}
\input{sections/05-Conclusion}

% \clearpage
\bibliography{main}
\bibliographystyle{unsrtnat}

\input{sections/06-Appendix}

\end{document}

%% file: math_commands.tex
\usepackage{amsmath,amsfonts,bm}

\def\eqref#1{equation~\ref{#1}}
\def\1{\bm{1}}

\DeclareMathAlphabet{\mathsfit}{\encodingdefault}{\sfdefault}{m}{sl}
\SetMathAlphabet{\mathsfit}{bold}{\encodingdefault}{\sfdefault}{bx}{n}

%% file: packages.tex
\usepackage[table]{xcolor}
\usepackage{titletoc}
\usepackage{tocloft}
\usepackage{lipsum}
\usepackage{url}
\usepackage{graphicx}
\usepackage{subcaption}
\usepackage{listings}
\usepackage[nolist]{acronym}
\usepackage{amsmath}
\usepackage[font={small}]{caption}
\usepackage[export]{adjustbox}
\usepackage{amssymb}
\usepackage{bm}
\usepackage{booktabs}
\usepackage{balance}
\usepackage{color}
\usepackage{dsfont}
\usepackage{esvect}
\usepackage{float}
\usepackage{graphicx}
\usepackage{import}
\usepackage{mathtools}
\usepackage{multirow}
\usepackage{multicol}
\usepackage{flushend}
\usepackage{outline}
\usepackage{paralist}
\usepackage{siunitx}
\usepackage{stackengine}
\usepackage{stmaryrd}
\usepackage{systeme}
\usepackage{soul}
\usepackage{url}
\usepackage{tikz}
\usetikzlibrary{tikzmark}
\usetikzlibrary{calc}
\usepackage[normalem]{ulem}
\let\labelindent\relax
\usepackage{enumitem}
\usepackage{algorithm2e}
\usepackage{scalerel}
\usepackage{makecell}

\definecolor{citecolor}{HTML}{0071bc}
\usepackage[pagebackref=true,breaklinks=true,urlcolor=blue,colorlinks,citecolor=citecolor,bookmarks=false]{hyperref}

%% file: macro.tex
\definecolor{codeblue}{rgb}{0.25,0.5,0.5}
\definecolor{myblue}{rgb}{0.92, 0.97, 0.85}

\definecolor{mygreen}{rgb}{0.92, 1.0, 0.92}
\definecolor{myred}{rgb}{1, 0.9, 0.9}
\definecolor{mygray}{gray}{0.95}
\definecolor{mydarkblue}{rgb}{0,0.08,1}
\definecolor{mydarkred}{rgb}{0.8,0.02,0.02}
\definecolor{mydarkorange}{rgb}{0.40,0.2,0.02}
\definecolor{mypurple}{RGB}{111,0,255}
\definecolor{mygold}{rgb}{0.75,0.6,0.12}
\definecolor{mydarkgray}{rgb}{0.66, 0.66, 0.66}
\definecolor{mydarkgreen}{rgb}{0.02,0.6,0.02}
\definecolor{mygray}{gray}{0.9}
\definecolor{keynotegreen}{rgb}{0.04,0.52,0}
\definecolor{keynoteyellow}{rgb}{1,0.68,0}
\definecolor{LightCyan}{rgb}{0.88,1,1}
\definecolor{tabfirst}{rgb}{1, 0.7, 0.7}
\definecolor{tabsecond}{rgb}{1, 0.85, 0.7} 
\definecolor{tabthird}{rgb}{1, 1, 0.7} 
\definecolor{rbtred}{rgb}{255, 0, 0}

\newcommand{\myparagraph}[1]{\vspace{0.1mm}\noindent\textbf{#1}}

\newcommand\blfootnote[1]{%
  \begingroup
  \renewcommand\thefootnote{}\footnote{#1}%
  \addtocounter{footnote}{-1}%
  \endgroup
}

\newlength\paramargin
\newlength\figmargin
\newlength\subfigmargin
\newlength\secmargin
\newlength\subsecmargin
\newlength\tabmargin
\newlength\eqmargin

\SetKwComment{Comment}{\color{green!50!black}\# }{}

\SetKwProg{Function}{def}{:}{}

\SetKwProg{For}{for}{:}{}
\SetKwProg{If}{if}{:}{}

\def\method{NaVILA\xspace}
\def\bench{VLN-CE-Isaac\xspace}

%% file: sections/00-Abstract.tex
\begin{abstract}
This paper proposes to solve the problem of Vision-and-Language Navigation with legged robots, which not only provides a flexible way for humans to command but also allows the robot to navigate through more challenging and cluttered scenes. However, it is non-trivial to translate human language instructions all the way to low-level leg joint actions. We propose \method, a 2-level framework that unifies a Vision-Language-Action model (VLA) with locomotion skills. Instead of directly predicting low-level actions from VLA, \method first generates mid-level actions with spatial information in the form of language, (e.g., ``moving forward 75cm''), which serves as an input for a visual locomotion RL policy for execution. \method substantially improves previous approaches on existing benchmarks. The same advantages are demonstrated in our newly developed benchmarks with IsaacLab, featuring more realistic scenes, low-level controls, and real-world robot experiments. 
% \url{https://navila-bot.github.io/} 
\end{abstract}

%% file: sections/01-Introduction.tex
% \vspace{-0.8em}
\section{Introduction}
% \vspace{-0.2em}

The ability to perform Vision-and-Language Navigation (VLN) has become a foundational component in modern robotics systems. With VLN, a robot is expected to navigate around unseen environments without a provided map following a language  instruction~\citep{anderson2018vision,wang2019reinforced,chaplot2020learning,chaplot2020object,chaplot2020neural,ramrakhya2022habitat}. This not only offers a better interface for humans, but also strengthen cross-scene generalization through languages. In this paper, we further extend the study of VLN with legged robots (e.g., quadruped or humanoid). Using legs instead of wheels allows robots to navigate in more challenging and cluttered scenarios. As the examples shown in Fig.~\ref{fig:teaser}, our robot can navigate through a messy laboratory space with narrow walkways, transition from room to room in a house, as well as tackle outdoor challenging environments such as uneven terrains with small rocks, holes, and troughs. 

% \vspace{-1.mm}

To translate language to action, the robot needs to reason about the input language, and perform closed-loop planning as well as low-level control. With the recent advancement in Large Language Models (LLMs) and Vision-Language Models (VLMs), several end-to-end Vision-Language-Action (VLA) systems have been developed~\citep{brohan2023rt,kim2024openvla,padalkar2024open}. These systems fine-tune a general-propose VLM with large-scale robot manipulation demonstrations to produce low-level actions. While unifying reasoning and execution in a single model is fascinating and shows encouraging results, it is worth diving deeper into the question: Is there a better way to represent actions beyond the quantized low-level commands? After all, LLMs and VLMs were primarily trained with natural language. Unifying reasoning and execution becomes challenging when we need to convert that reasoning into precise, non-verbal actions.

% \vspace{-1.mm}

Inspired by the recent progress on VLM~\citep{chen2024spatialvlm,cheng2024spatialrgpt} for spatial location and distance reasoning, we propose \textbf{\method}, a two-level framework for legged robot VLN: A VLM is fine-tuned to output a \textbf{mid-level action} (VLA) in the form of language such as ``turn right 30 degrees'', and a low-level visual locomotion policy is trained to follow this instruction for execution. The mid-level action output of the VLA conveys the location and direction information without the low-level commands. The advantages of this framework are three-fold: (i) By decoupling low-level execution from VLAs, the same VLA can be applied across different robots by swapping the low-level policy; (ii) Representing actions as mid-level language instructions enables VLA training with diverse data sources, including real human videos and reasoning QA tasks. This enhances reasoning capabilities without overfitting outputs to specific low-level commands and can leverage real-world data for generalization; (iii) \method operates on two distinct timescales: the VLA, typically a large and computationally intensive model, runs at a lower frequency, providing high-level navigation commands; while the locomotion policy operates in real-time. This dual-frequency approach allows the locomotion policy to handle sophisticated obstacle avoidance and increases overall robustness.

% \vspace{-1.mm}

To train the VLA, we demonstrate how to (i) integrate historical context and current observations in VLN within existing VLM frameworks, (ii) create a specialized navigation prompt tailored for VLN tasks, (iii) utilize real-world data from YouTube human touring videos to improve navigation in continuous environments, and (iv) introduce a carefully curated dataset blend designed to enhance VLN generalizability. These strategies allow us to fine-tune a general-purpose image-based VLM into a navigation-focused agent while simultaneously training it on general vision-language datasets, thereby maintaining its broad generalization capabilities. Moreover, this is the first work to show that direct training on human videos improves navigation in continuous environments.

To train robust locomotion skills, we employ a single-stage approach to learn vision-based locomotion policy. We construct a height map from raw LiDAR point clouds and introduce randomization to bridge the sim-to-real gap. This controller takes the output from our VLA model, converts it into command velocities, and tracks these velocities by controlling the positions of the joints. This end-to-end approach enables the training of visual locomotion skills that are both robust and safe, facilitating deployment in real-world, challenging environments (e.g., strong sunlight or near certain transparent surfaces).

% \vspace{-1.mm}

In our experiments, we show that our VLA significantly outperforms the state-of-the-arts on classic VLN benchmarks, with over 17\% improvement in success rate. Additionally, our single-stage locomotion policy outperforms previous policy distillation-based methods by a substantial margin. To better simulate the challenges of locomotion navigation in VLN, we introduce a new benchmark, \bench, using Isaac Sim. This benchmark considers detailed robotic joint movements and interactions with environments, which prior VLN works have not explored. In our \bench experiments, our vision-based policy outperforms the blind policy by a significant margin, showing a 14\% improvement in success rate. We also demonstrate that our VLA can be deployed across different robots (Unitree Go2, Unitree H1, Booster T1), each using distinct locomotion skills. Finally, we deploy \method in the real world, exhibiting impressive robustness and achieving an 88\% success rate on 25 instructions, including a 75\% success rate on complex instructions across diverse scenes.

%% file: sections/03-Method2.tex
\begin{figure*}[t]
  \centering
  \includegraphics[width=1\textwidth]{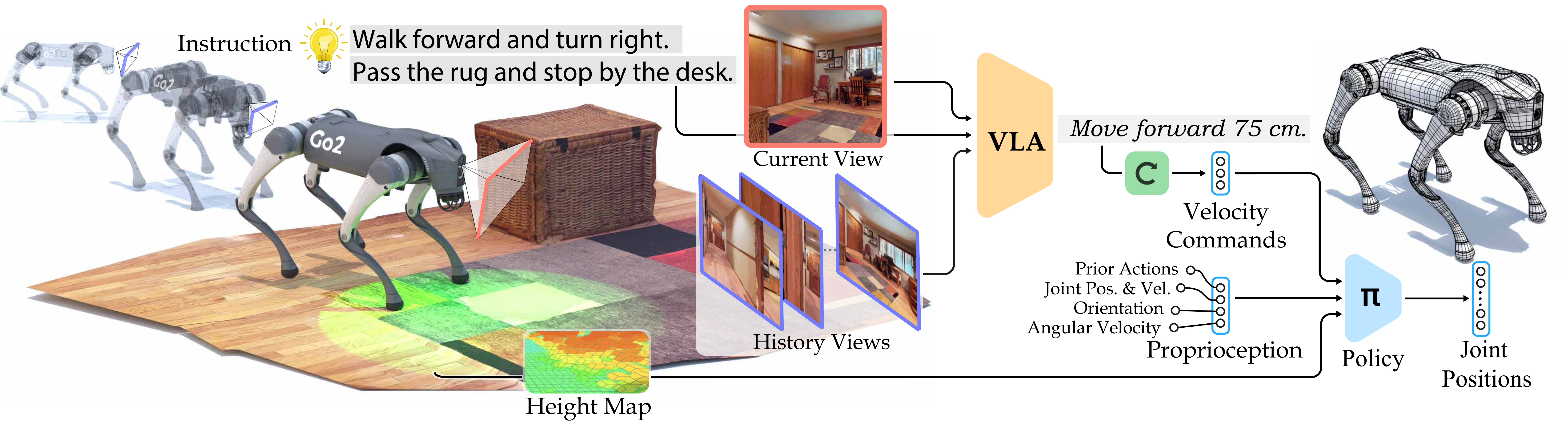}
  \caption{\method is a two-level framework combining high-level visual language understanding with low-level locomotion control. Our VLA model processes single-view images to produce mid-level actions in natural language, which are then converted into precise joint movements by an advanced low-level locomotion policy. This integration allows for strong generalization and adaptability across different real-world environments, and can operate the robot in real-time.}
  \label{fig:overview}
  \vspace{-15pt}
\end{figure*}

\vspace{-1.5em}
\section{Method}
\label{sec:method}
\vspace{-0.2em}
% \method integrates high-level visual language understanding and action with low-level locomotion control (Fig.~\ref{fig:overview}). Specifically, \method employs a VLM that processes single-view images to generate waypoint instructions in natural language. These instructions are then interpreted by a low-level locomotion policy, which translates them into precise joint movements for real-time robot control. The synergy between the VLM's high-level reasoning and the locomotion policy's execution capabilities enables \method to demonstrate remarkable generalization and adaptability across diverse real-world environments. In the following sections, we begin by describing how we tame VLMs for high-level VLN in Sec.~\ref{sec:method:high}, followed by an overview of our robot configuration and low-level locomotion policy in Sec.~\ref{sec:method:low}.

\method integrates high-level visual language understanding with low-level locomotion control (Fig.\ref{fig:overview}). It employs a VLM to process single-view images and generate waypoint instructions in natural language, which a locomotion policy translates into precise joint movements for real-time robot control. The synergy between the VLM’s reasoning and the locomotion policy’s execution enables \method to generalize across diverse environments. We first describe how we tame VLMs for high-level VLN in Sec.\ref{sec:method:high}, then outline our robot configuration and locomotion policy in Sec.~\ref{sec:method:low}.

\subsection{Taming VLMs for Vision Language Navigation}
\label{sec:method:high}
VLN requires processing video inputs as observations. A common approach to handling video inputs in VLMs is through video encoders~\cite{zhang2024navid}. However, recent progress in VLMs has largely been driven by the availability of image-text data. While there have been efforts to extend this success to video encoders, the lack of large, high-quality video-text datasets has limited their pre-training. To address this challenge, we opt for image-based vision-language models in our approach. These models exhibit stronger generalization abilities and possess broader knowledge, making them more suitable for tackling the generalization challenges in VLN. Specifically, we built our approach upon VILA~\citep{lin2024vila,wu2024vila,fang2024vila,xue2024longvila,ye2024x,huang2025lita,liu2024nvila}, a family of efficient VLMs for both understanding and generation. VILA's pre-training has proven particularly effective for multi-image reasoning, making it especially suitable for VLN tasks where understanding sequential image relationships is critical.

\myparagraph{VILA Preliminary.}
VILA consists of three main components: a vision encoder, a projector, and an LLM. The vision encoder processes the input images, converting them into a sequence of visual tokens. These tokens are then downsampled and mapped into the language domain via an MLP projector. Afterward, the projected tokens, along with text tokens, are sent to the LLM for auto-regressive generation. When handling videos, VILA uniformly sampled frames at regular intervals. It puts all the frame information before any text. A typical prompt for describing a video might look like ``\texttt{$\langle$frame3$\rangle$$\langle$frame6$\rangle$$\langle$frame9$\rangle$...Tell me about this video.}'' Notably, with sequence parallel training~\citep{xue2024longvila}, VILA can include frames up to 1024. VILA undergoes a 3-stage training process: first, it pre-trains a connector between the frozen LLM and vision backbones using alignment data~\citep{liu2023visual}; then it pre-trains both the connector and the LLM using text-image interleaved corpus~\citep{coyo,zhu2024multimodal}; and finally, it fine-tunes all modules (vision encoder, connector, LLM) with instruction tuning data~\citep{liu2023visual,liu2024improved}.

\myparagraph{Navigation Prompts.}
In vision-language navigation tasks, images from different time steps serve two distinct purposes. The image at time step $t$ represents the current observation, which is crucial for a VLN agent to make immediate decisions (e.g., turning right at an intersection or stopping when the goal is reached). On the other hand, frames before time step $t$ are historical frames that function as a memory bank, helping the agent track overall progress (e.g., remembering the starting location, reasoning about places already visited and planning the next step). Uniformly sampling frames at regular intervals, as done in VILA, is not ideal because it doesn't differentiate between these two types of representations. Therefore, we first extract the most recent frame $t$ as the current observation and then uniformly sample frames from the preceding $t-1$ frames, ensuring the first frame is always included. Additionally, since current and historical observations serve different roles, we distinguish them in our task prompt using textual cues like \texttt{a video of historical observations:} for memory frames and \texttt{current observation:} for the latest frame. Unlike~\citep{zhang2024navid}, we avoid introducing additional special tokens that could complicate the LLM's learning process. Instead, we adhere to our design principle of keeping both the input and output of LLM in the language domain to fully leverage the reasoning capabilities of the pre-trained LLM. By integrating these tokens for historical and current observations with the navigation instruction, we construct a navigation task prompt, as shown in Fig.~\ref{fig:overview}.

\begin{figure*}[t]
  \centering
  \includegraphics[width=0.95\textwidth]{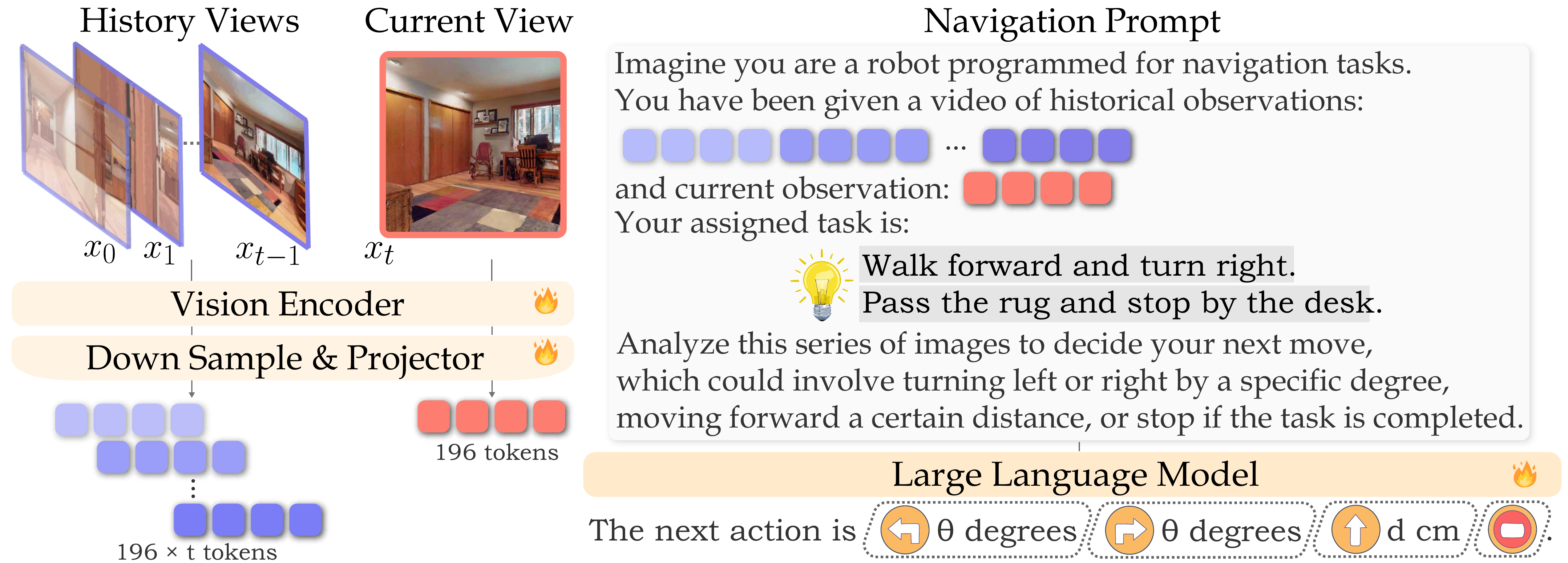}
  \caption{Overview of our VLA framework. We denote the purple blocks (\protect\scalerel*{\includegraphics{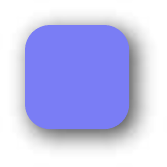}}{A}) as memory tokens sampled from historical frames, and the red blocks (\protect\scalerel*{\includegraphics{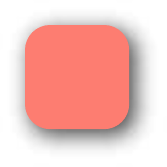}}{A}) as the current observation tokens. \protect\scalerel*{\includegraphics{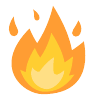}}{A} denotes trainable parameters. In our experiments, we tested configurations with 8 to 64 frames for $t$.}
  \label{fig:vlm-arch}
\vspace{-15pt}
\end{figure*}

\myparagraph{Learning from Human Videos.}
Recent studies~\citep{guhur2021airbert,majumdar2020improving,lin2023learning} have shown that collecting trajectory-instruction pairs from human videos can enhance navigation capabilities. However, prior work has been limited to discrete navigation settings and has mainly used real videos for pre-training to reduce domain gaps or improve landmark understanding, rather than for directly training navigation models. Extending this approach to continuous settings presents a significant challenge due to the difficulty of obtaining continuous action labels. Recent advances in metric-pose estimation in the wild have now made this feasible, enabling us to extract spatial understanding from human videos and train navigation models directly.

\begin{figure}[t]
  \centering
  \includegraphics[width=0.49\textwidth]{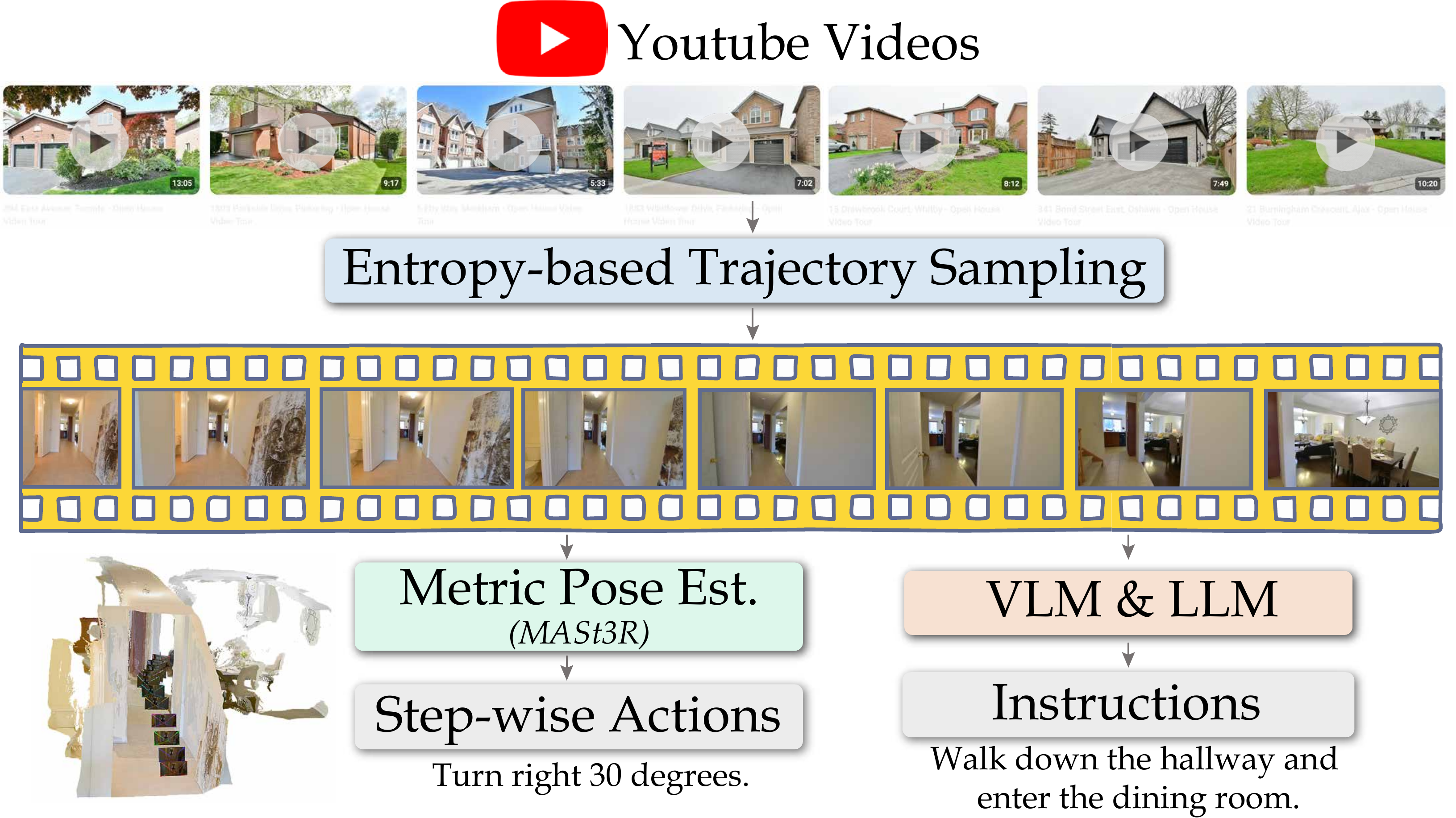}
  \caption{Data pipeline for transforming human touring videos in the wild into pairwise navigation data within a continuous environment. We begin by processing the videos into meaningful trajectories through entropy-based sampling~\citep{lin2023learning}. Then we extract step-wise actions through metric camera pose estimation~\citep{leroy2024grounding}, and utilize VLM~\citep{lin2024vila} and LLM~\citep{openai2024gpt4o} to generate instructions.}
  \label{fig:data-pipeline}
\vspace{-20pt}
\end{figure}

Our data pipeline, shown in Fig.~\ref{fig:data-pipeline}, starts with 2K egocentric touring videos from YouTube, which provide a rich source of real-world data to learn robot navigation from human behavior. We process these videos into 20K diverse and representative trajectories using entropy-based sampling~\citep{lin2023learning}. Next, we estimate camera poses using MASt3R~\citep{leroy2024grounding} to extract step-by-step actions, and we generate natural language instructions for each trajectory using VLM-based~\citep{lin2024vila} captioning followed by LLM~\citep{openai2024gpt4o} rephrasing. This approach allows us to leverage human demonstrations for continuous navigation, a capability that was previously non-trivial to achieve.

% First, we incorporate navigational data from real human touring videos. 
\myparagraph{Supervised Fine-tuning Data Blend.}
Effective Supervised Fine-tuning (SFT) data is crucial for developing a robust vision-language action model. The model should specialize in embodied tasks while avoiding overfitting to specific actions. It should also generalize well to real-world scenarios while retaining broad-world knowledge. Thanks to NaVILA's modular framework, which offers exceptional scalability and adaptability, integrating diverse data sources into our pipeline is straightforward. This flexibility allows us to enhance generalizability for navigation. Our SFT data blend is designed from four perspectives: (1) Navigational data from real videos, (2) Navigational data from simulations, (3) Auxiliary navigational data, and (4) General VQA datasets.

For simulated navigational data, the available VLN datasets in continuous environments are limited, with only R2R-CE~\citep{r2r} and RxR-CE~\citep{rxr} providing sparse path points converted from discrete VLN versions. We leverage both datasets within the Habitat simulator, using a shortest path follower to generate action sequences along the geodesic shortest path. This results in step-wise navigation videos, where each sample comprises a $(t+1)$-frame video and the corresponding oracle action at time step $t$. To encourage the LLM to generate continuous value labels for distances and angles, we merge consecutive actions (e.g., combining two forward 25 cm steps into a single forward 50 cm step), with a maximum of three consecutive actions. This merging process not only reduces dataset size for more efficient processing but also introduces greater diversity in actions, mitigating overfitting. Additionally, to address label imbalance—particularly the underrepresentation of the stop action—we apply a rebalancing technique for a more even distribution. All navigation-specific data undergo the previously described frame extraction strategy and are paired with navigation task prompts.

To further improve scene understanding and address the limited instructions in R2R-CE and RxR-CE, we incorporate auxiliary navigational datasets. Following~\citep{zhang2024navid}, we use augmented instructions from EnvDrop~\citep{tan2019learning} and introduce an auxiliary task of navigation trajectory summarization. Given a trajectory video, we sample frames by retaining the first frame and uniformly selecting historical frames, using the annotated instructions as labels. The LLM is then tasked with describing the robot’s trajectory based on these frames. To further enhance spatial scene understanding, we integrate the ScanQA~\citep{azuma2022scanqa} dataset, which features real-world 3D scan QA pairs with human-edited questions and free-form answers grounded in 3D objects. For training, we use multi-view RGB images from the raw scans to support this task.

Finally, to maintain the model’s general capabilities, we incorporate general VQA datasets from~\citep{liu2024improved, chen2023sharegpt4v, maaz2023videochatgpt}. This comprehensive dataset design ensures that \method can generalize effectively to novel scenes and real-world environments.

\myparagraph{Training and Inference Paradigm.}
Our training process begins with the stage two model of VILA, which has already undergone visual language corpus pre-training. We then apply our SFT data blend to train the entire VLM for one epoch, following standard practices. During this training, all three components—vision encoder, connector, and LLM—are unfrozen.
For the inference phase, we implement a regular expression parser~\citep{kearns1991extending}, to extract action types (such as forward or turn left) and their corresponding arguments (like specific distance or angles) from the LLM output. This method has demonstrated effectiveness in both simulated environments and real-world experiments, where we empirically found that all actions throughout all experiments are successfully matched and mapped.

\subsection{Visual Locomotion Policy}

\label{sec:method:low}
In this section, we begin with a brief overview of the Go2 robot dog, the experimental platform used in this work. Next, we describe the development of the end-to-end vision-based control policy, which interprets high-level language navigation commands from the VLM and converts them into precise joint movements. This control policy is trained in the Isaac Sim simulator using Isaac Lab~\citep{mittal2023orbit} and then directly deployed to the real-world robot.

\begin{figure}[t]
  \centering
  \includegraphics[width=.49\textwidth]{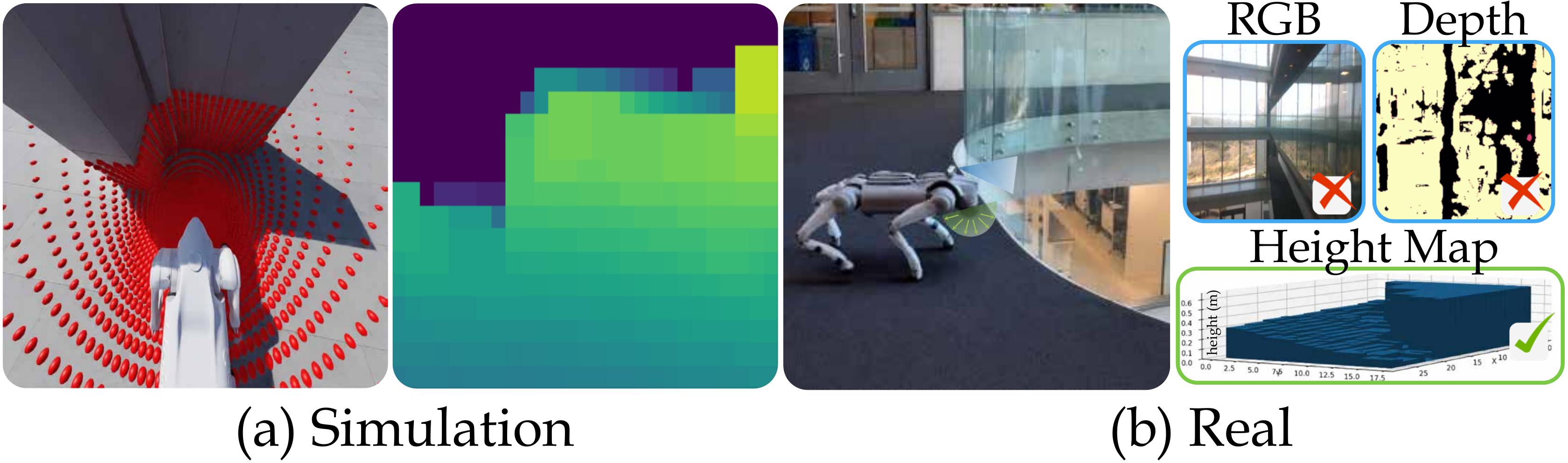}
  \vspace{-10pt}
  \caption{Height map reconstruction from point cloud. (\textbf{a}) Go2 robot follows velocity commands while avoiding obstacles in simulation. Red dots show LiDAR points raycasting from the sensor center to the terrain mesh. The right image shows a preprocessed height map with values clipped to sensor constraints; darker colors indicate higher heights. (\textbf{b}) Safe locomotion near glass. The top-down height map detects glass surfaces where depth and RGB images fail.}
  \label{fig:lidar_method}
  \vspace{-15pt}
\end{figure}

\myparagraph{Go2 Robot.}
As shown in Fig.~\ref{fig:lidar_method}, the robot is equipped with a LiDAR sensor mounted at the base of its head, broadcasting point clouds at a frequency of 15Hz. The robot features 18 degrees of freedom (DoFs), comprising 6 DoFs for its base and 3 DoFs for each of its four legs. In the policy training process, we left the 6 DoFs on the base unconstrained so that the policy only controls the 12 joint motors on the legs.

\myparagraph{Interpreting High-level Commands.}
As in our formulation, VLM outputs a fixed set of actionable words, such as \{move forward, turn left, turn right, stop\}, we casts these instructions to fixed command velocities \{$0.5 \,\si{\meter\per\second}$, $\frac{\pi}{6} \, \si{\radian\per\second}$, $-\frac{\pi}{6} \, \si{\radian\per\second}$, 0\} and execute with corresponding time durations to align with the specific VLM value.
% Need to describe how to convert the text instructions into velocities.

\myparagraph{Low-level Action and Observation Space.}
The action space $\mathbf{a}$ of the control policy is defined as the desired joint position $q^d \in \mathbb{R}^{12}$, which is converted into torque input for the simulator using the stiffness and dampness. We adopt the PPO algorithm~\cite{schulman2017proximal} to train the policy. During training, the critic observes the privileged environment and generates a value function to update the actor,  while the actor only receives sensor data available in the real world. The observation space of the critic $\mathbf{o}^{c}$ contains the proprioception and velocity command at the current time step $t$ and a privileged terrain height scan around the robot. The proprioceptive data includes robot linear and angular velocity, orientation, joint positions, joint velocities, and the previous action. In the actor's observation space, $\mathbf{o}^{a}$, linear velocity is excluded, as it is unavailable in the real world, and instead, a history of proprioceptive data is used to infer this information implicitly. The robot perceives the surrounding terrain using a height map from the LiDAR sensor.

\myparagraph{Incorporating Height Map from LiDAR Point Cloud.}
Given LiDAR's superior ability to detect transparent objects and robust performance under strong sunlight,  we chose the manufacturer-provided LiDAR as the primary sensor for perceiving the robot's surroundings and ensuring safe navigation. The Unitree L1 generates point clouds with a wide field of view of $360^{\circ}\times90^{\circ}$, from which we create a 2.5D height map based on the parameters listed in the Supplementary. For each voxel grid, the lowest value within the range is selected, and a maximum filter is then applied over the last 5 lidar point clouds to smooth the resulting height map.

\myparagraph{Training.} 
Different from most existing works~\citep{lee2020learning, miki2022learning, margolis2022rapid, lee2024learning} that utilize the two-stage teacher-student training paradigm, we adopt a single-stage manner to train the locomotion policy. Compared to two-stage training, single-stage RL is more time-efficient as it eliminates the need for policy distillation. Additionally, the policy interacts directly with the environment, allowing it to explore and potentially discover novel strategies. With the support of ray-casting in Isaac Lab, our vision-based RL policy training achieves a high throughput over 60K FPS on an RTX 4090 GPU.

%% file: sections/04-Experiments.tex
\input{tables/r2r_rxr}

\input{tables/rxr_cross}

\section{Experiments}

We conduct experiments to answer the following questions: (1) How does our VLA's performance compare to state-of-the-art methods in VLN-CE benchmarks and general spatial scene understanding tasks? (Sec.~\ref{exp:vlm}) (2) How does the performance of our single-stage visual locomotion policy compare to policy distillation-based approaches? (Sec.~\ref{exp:loco}) (3) How to evaluate locomotion navigation in simulators, and how effective and flexible is \method in these scenarios? (Sec.~\ref{exp:loco-vlnce}) (4) Can \method pipeline be successfully deployed in real robot VLN experiments? (Sec.~\ref{exp:real})

\subsection{High-level VLA Performance}
\label{exp:vlm}
\vspace{5pt}
\myparagraph{VLN-CE Benchmarks.}
We evaluate our VLA on the VLN-CE benchmarks, which provide continuous environments for executing navigational actions in reconstructed photorealistic indoor scenes. We focus on the val-unseen split in both R2R (Room-to-Room) and RxR (Room-across-Room) datasets within VLN-CE, as these are the two most recognized benchmarks in VLN. We employ the following widely used evaluation metrics for VLN tasks: Navigation Error (NE), Oracle Success Rate (OS), Success Rate (SR), Success-weighted Path Length (SPL), and normalize dynamic time wrapping (nDTW). We show results in Table~\ref{tab:r2r_rxr}, where \method significantly surpasses all baselines that do not rely on simulator pre-trained waypoint predictors in both benchmarks using a single model. Notably, this also marks the first time a VLN agent, trained solely on single-view RGB input, achieves comparable or superior results to models that use panoramic views, odometry, or simulator-pretrained waypoint predictors. This suggests that \method's strong generalization capabilities can effectively compensate for the limited observations in RGB views or sensors.

To evaluate the cross-dataset performance, we follow \citep{zhang2024navid} by training \method exclusively on R2R samples, while leaving out the RxR training set. We then evaluate its zero-shot performance on the RxR Val-Unseen split. As shown in Table~\ref{tab:rxr_cross}, our method significantly outperforms NaVid, the current state-of-the-art model, with a substantial 10\% improvement in SR.

\myparagraph{Spatial Scene Understanding Benchmarks.}
As a general navigation agent, robust spatial scene understanding (e.g., object localization, referring, and spatial reasoning) is crucial. To evaluate \method's capabilities in scene understanding, we conduct evaluations on the ScanQA Validation benchmark, a widely used dataset for 3D Question Answering. ScanQA is based on real-world scans, and we use multi-view images from these scans as input to query \method for answers.
As shown in Table~\ref{tab:scanqa}, \method significantly outperforms the previous state-of-the-art model, NaviLLM~\citep{zheng2024towards}, by a substantial margin (20 points higher on the CIDEr score). Moreover, when using 64 frames, NaVILA's performance demonstrates superior performance compared to state-of-the-art 3D-based large multi-modal models~\citep{fu2024scene, huang2023embodied}. This is particularly noteworthy as these other models require either 3D scans or RGBD data with camera poses as inputs, while our method achieves better results with less observation.

\input{tables/scanqa}

\subsection{Low-level RL Policy Performance}
\label{exp:loco}

To highlight the advantages of our RL policy over policy distillation-based approaches, we compared it to Regularized Online Adaptation (ROA) \citep{fu2022deep}. In ROA training, the model first learns a privileged encoder that processes height scan points and other privileged observations. This privileged encoder then supervises an adaptation encoder, which takes the same 2.5D heightmap as our low-level policy as input. We evaluated both approaches using three metrics: linear velocity error, angular velocity error, and collision rate. The first two metrics assess how accurately the policy follows velocity commands, while the third measures the model’s obstacle avoidance capabilities. As shown in Table~\ref{tab:low_level}, our low-level policy outperforms ROA in all three metrics, particularly achieving a significantly lower collision rate, demonstrating the effectiveness of our training approach.

\subsection{Legged Robot Navigation Performance in Simulation}
\label{exp:loco-vlnce}
\myparagraph{High-fidelity \bench Benchmark.}
Currently, there are no VLN-CE benchmarks tailored specifically for legged robots. Existing benchmarks~\citep{r2r,rxr} for vision-language navigation are based on the Habitat~\citep{savva2019habitat} simulator, which focuses on high-level planning without addressing precise low-level robotic control. For instance, agents in Habitat can navigate through narrow gaps, such as a 10 cm space between two sofas, which is impractical for legged robots like quadrupeds or humanoids. To overcome this limitation, we introduce a new benchmark \bench built on Isaac Sim. Isaac Sim's high-fidelity simulation captures detailed robotic joint movements and interactions with the environment, enabling comprehensive evaluations of the entire navigation pipeline, from high-level planning to precise robotic execution. We incorporate the same scenes from R2R, with robots deployed in the environment, as shown in Fig.~\ref{fig:issac-vis}. From the 1,839 trajectories in the R2R Val-Unseen split, we select 1,077 traversable trajectories with high-quality meshes to ensure realistic navigation scenarios. For consistency, we evaluate performance using the same metrics as prior work.

\input{tables/r2r_isaac}
\input{tables/low_level}

Notably, \bench is compatible with a variety of robotic platforms. To demonstrate this flexibility, we test our \method model on a Unitree Go2 robot and also a Unitree H1 robot within the benchmark. To highlight the effectiveness of the vision-based policy, we compare it against a proprioception-only (\textit{blind}) policy. 
As shown in Table~\ref{tab:r2r_isaac}, the vision-based policy outperforms the blind policy by 14\% in Success Rate in Go2 settings and 21\% in H1 settings, owing to its superior obstacle avoidance capability. We also compare NaVILAs with a baseline using Oracle’s low-level policy (assuming perfect command execution without realistic physics). Results show a 15\% lower success rate on the Go2 setup and a 27\% lower success rate on the H1 setup when Oracle policy is not presented. These performance gaps highlight the increased challenges and realism introduced by our benchmark. Additionally, we also observe that the success rate of \method on the H1 robot is significantly lower than on the Go2, which is expected due to the larger size of the humanoid robot.

% To further demonstrate the effectiveness of our proposed low-level locomotion policy, we compared it with two-stage methods \citep{kumarrma2021} and results are reported in ... Our method outperforms in both linear and angular velocity error, and achieves a much lower collision rate.

\subsection{Real World Evaluation}
\label{exp:real}
\begin{figure*}[!th]
  \centering
  \includegraphics[width=1.0\textwidth]{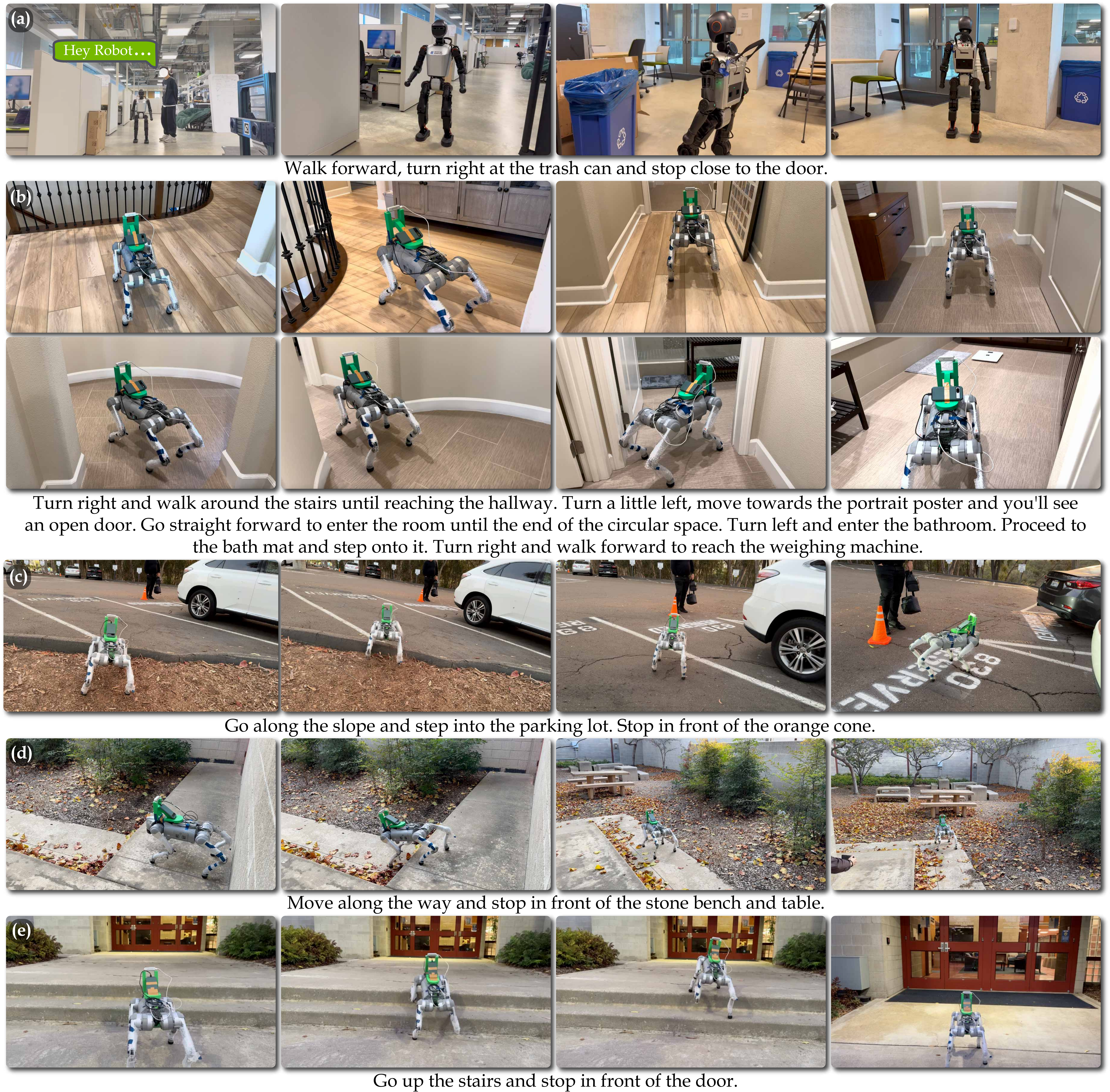}
\vspace{-1.6em}
  \caption{Qualitative results from the real-world deployment of \method. (a) We integrate speech recognition~\cite{radford2023robust} into \method, allowing a human to control the robot using voice commands that begin with \texttt{"Hey Robot!"}. (b) The robot successfully handles long-horizon navigation tasks. Given a lengthy instruction, it moves through different areas of the house and stops at the specified goal. (c), (d), and (e) The robot demonstrates its ability to navigate through obstacles, traverse challenging terrains, and climb up and down stairs.}
  \label{fig:exp-real}
  \vspace{-20pt}
\end{figure*}

We then conduct experiments in the real world, using 25 instructions, each repeated three times, covering both simple and complex tasks across three types of environments: \texttt{Workspace}, \texttt{Home}, and \texttt{Outdoor} open environments. Simple instructions consist of one or two navigation commands, where the robot does not need to navigate between rooms (e.g., ``Go to the chair and stop''). In contrast, complex instructions involve three or more commands, requiring the robot to traverse multiple rooms or landmarks (e.g., ``Walk out of the room, turn right, enter the room in front of you, and stop at the table''). We use standard metrics (SR and NE) and compare \method against GPT-4o, a state-of-the-art VLM known for its strong generalizability. As shown in Table~\ref{tab:real_eval}, \method significantly outperforms GPT-4o in both SR and NE. We also ablate the effectiveness of adding human videos, and the results show that with the help of human videos, the model can generalize better to outdoor scenes and achieve higher success rates across all environments. To demonstrate the flexibility of our two-level approach, we also evaluated it on a Booster Dynamics T1 humanoid robot, using the same VLA model without any retraining. Despite variations such as changes in camera height and camera view angle, \method consistently outperformes the baselines, highlighting the strong generalization capabilities of our model.
% Note that we also conduct experiments using the G1 humanoid robot and achieved these results using the same VLA model without any retraining for the humanoid robot.
Our qualitative results are presented in Fig.\ref{fig:teaser} and Fig.\ref{fig:exp-real}. In Fig.~\ref{fig:exp-real}, we demonstrate integration with speech recognition, enabling voice-controlled navigation through our framework. These results highlight the effectiveness of \method in bridging the gap between vision-language understanding and real-world navigation tasks.

% Due to the constraints of low-level policy, we tested with simple and complex instructions in the \texttt{Workspace} and simple instructions in the \texttt{Outdoor} environment.

\input{tables/real_eval}

%% file: tables/r2r_rxr.tex
\begin{table*}[t]
    \small
    \centering
    \caption{Comparison with state-of-the-art methods on the Val-Unseen split of R2R-CE~\citep{r2r} and RxR-CE~\citep{rxr}. $^{*}$ indicates methods using the waypoint predictor from~\citet{hong2022bridging}. \method outperforms all methods that do not rely on simulator pre-trained waypoint predictors, even when those methods leverage additional inputs such as depth, panoramic views, and odometry.}
    % \vspace{-5pt}
    \setlength{\tabcolsep}{7.0pt}
    \scalebox{0.97}{
{\fontsize{8pt}{9pt}\selectfont
\begin{tabular}{lcccclcccclcccc}
\toprule
& \multicolumn{4}{c}{Observation} & & \multicolumn{4}{c}{R2R Val-Unseen} & & \multicolumn{4}{c}{RxR Val-Unseen} \\
\cmidrule(lr){2-5} \cmidrule(lr){7-10} \cmidrule(lr){12-15}
& S.RGB & Pano. & Depth & Odo. & & NE $\downarrow$ & OS $\uparrow$ & SR $\uparrow$ & SPL $\uparrow$ & & NE $\downarrow$ & SR $\uparrow$ & SPL $\uparrow$ & nDTW $\uparrow$ \\
\midrule
HPN+DN$^{*}$~\citep{krantz2021waypoint} &  & \checkmark & \checkmark & \checkmark & & 6.31 & 40.0 & 36.0 & 34.0 & & - & - & - & - \\
CMA$^{*}$~\citep{hong2022bridging} & & \checkmark & \checkmark & \checkmark & & 6.20 & 52.0 & 41.0 & 36.0 & & 8.76 & 26.5 & 22.1 & 47.0 \\
VLN$\circlearrowright$BERT$^{*}$~\citep{hong2022bridging} & & \checkmark & \checkmark & \checkmark & & 5.74 & 53.0 & 44.0 & 39.0 & & 8.98 & 27.0 & 22.6 & 46.7 \\
Sim2Sim$^{*}$~\citep{krantz2022sim} & & \checkmark  & \checkmark & \checkmark & & 6.07 & 52.0 & 43.0 & 36.0 & & - & - & - & - \\
GridMM$^{*}$~\citep{wang2023gridmm} & & \checkmark  & \checkmark & \checkmark & & 5.11 & 61.0 & 49.0 & 41.0 & & - & - & - & - \\
Ego$^{2}$-Map$^{*}$~\citep{hong2023learning} & & \checkmark  & \checkmark & \checkmark & & 5.54 & 56.0 & 47.0 & 41.0 & & - & - & - & - \\
DreamWalker$^{*}$~\citep{wang2023dreamwalker} & & \checkmark  & \checkmark & \checkmark & & 5.53 & 59.0 & 49.0 & 44.0 & & - & - & - & - \\
Reborn$^{*}$~\citep{an20221st} & & \checkmark  & \checkmark & \checkmark & & 5.40 & 57.0 & 50.0 & 46.0 & & 5.98 & 48.6 & 42.0 & 63.3 \\
ETPNav$^{*}$~\citep{an2024etpnav} & & \checkmark  & \checkmark & \checkmark & & 4.71 & 65.0 & 57.0 & 49.0 & & 5.64 & 54.7 & 44.8 & 61.9 \\
HNR$^{*}$~\citep{wang2024lookahead} & & \checkmark  & \checkmark & \checkmark & & 4.42 & 67.0 & 61.0 & 51.0 & & 5.50 & 56.3 & 46.7 & 63.5 \\
BEVBert$^{*}$~\citep{an2023bevbert} & & \checkmark  & \checkmark & \checkmark & & 4.57 & 67.0 & 59.0 & 50.0 & & 4.00 & 68.5 & - & 69.6 \\
HAMT+ScaleVLN$^{*}$~\citep{wang2023scaling} & & \checkmark  & \checkmark & \checkmark & & 4.80 & - & 55.0 & 51.0 & & - & - & - & - \\
\midrule
AG-CMTP~\citep{chen2021topological} & & \checkmark  & \checkmark & \checkmark & & 7.90 & 39.0 & 23.0 & 19.0 & & - & - & - & - \\
R2R-CMTP~\citep{chen2021topological} & & \checkmark  & \checkmark & \checkmark & & 7.90 & 38.0 & 26.0 & 22.0 & & - & - & - & - \\
LAW~\citep{raychaudhuri2021language} & \checkmark & & \checkmark & \checkmark & & 6.83 & 44.0 & 35.0 & 31.0 & & 10.90 & 8.0 & 8.0 & 38.0 \\
CM2~\citep{georgakis2022cross} & \checkmark & & \checkmark & \checkmark & & 7.02 & 41.0 & 34.0 & 27.0 & & - & - & - & - \\
WS-MGMap~\citep{chen2022weakly} & \checkmark & & \checkmark & \checkmark & & 6.28 & 47.0 & 38.0 & 34.0 & & - & - & - & - \\
AO-Planner~\citep{chen2024affordances} & & \checkmark & \checkmark & & & 5.55 & 59.0 & 47.0 & 33.0 & & 7.06 & 43.3 & 30.5 & 50.1 \\
Seq2Seq~\citep{krantz2020beyond} & \checkmark & & \checkmark & & & 7.77 & 37.0 & 25.0 & 22.0 & & 12.10 & 13.9 & 11.9 & 30.8 \\
CMA~\citep{krantz2020beyond} & \checkmark & & \checkmark & & & 7.37 & 40.0 & 32.0 & 30.0 & & - & - & - & - \\
RGB-Seq2Seq~\citep{krantz2020beyond} & \checkmark & & & & & 10.10 & 8.0 & 0.0 & 0.0 & & - & - & - & - \\
RGB-CMA~\citep{krantz2020beyond} & \checkmark & & & & &  9.55 & 10.0 & 5.0 & 4.0 & & - & - & - & - \\
NaVid~\citep{zhang2024navid} & \checkmark & & & & &  5.47 & 49.0 & 37.0 & 35.0 & & - & - & - & - \\
\rowcolor{myblue}
\textbf{\method} & \checkmark  & & & & & \bf5.22 & \bf62.5 & \bf54.0 & \bf49.0 & & \bf6.77 & \bf49.3 & \bf44.0 & \bf58.8 \\
\bottomrule
\end{tabular}}}
\vspace{-15pt}
\label{tab:r2r_rxr}
\end{table*}

%% file: tables/rxr_cross.tex
\begin{table}[t]
    \small
    \centering
    \caption{Cross-dataset performance on the RxR-CE~\citep{rxr} Val-Unseen split. All results are obtained without training on the RxR-CE training set. \method significantly outperforms NaVid~\citep{zhang2024navid}, the current single-view state-of-the-art.}
    % \vspace{-5pt}
    \setlength{\tabcolsep}{3.pt}
    \scalebox{0.97}{
{\fontsize{8pt}{9pt}\selectfont
\begin{tabular}{lccclcccc}
\toprule
& \multicolumn{3}{c}{Observation} & & \multicolumn{4}{c}{RxR Val-Unseen}\\
\cmidrule(lr){2-4} \cmidrule(lr){6-9}
& S.RGB & Depth & Odo. & & NE $\downarrow$ & OS $\uparrow$ & SR $\uparrow$ & SPL $\uparrow$ \\
\midrule
LAW~\citep{raychaudhuri2021language} & \checkmark & \checkmark & \checkmark & & 10.87 & 21.0 & 8.0 & 8.0 \\
CM2~\citep{georgakis2022cross} & \checkmark & \checkmark & \checkmark & & 8.98 & 25.3 & 14.4 & 9.2 \\
WS-MGMap~\citep{chen2022weakly} & \checkmark & \checkmark & \checkmark & & 9.83 & 29.8 & 15.0 & 12.1 \\
 Seq2Seq~\citep{krantz2020beyond} & \checkmark & \checkmark & &
 & 11.8 & 5.02 & 3.51 & 3.43 \\
 CMA~\citep{krantz2020beyond} & \checkmark & \checkmark & & 
 & 11.7 & 10.7 & 4.41 & 2.47 \\
 \midrule
 RGB-Seq2Seq~\citep{zhang2024navid} & \checkmark & & & &  11.2 & 12.2 & 0.0 & 0.0 \\
 RGB-CMA~\citep{zhang2024navid} & \checkmark & & & 
 & 9.55 & 14.8 & 0.0 & 0.0 \\
 A$^{2}$NAV~\citep{chen2023a2nav} & \checkmark & & & 
 & - & - & 16.8 & 6.3 \\
 NaVid~\citep{zhang2024navid} & \checkmark & & & 
 & \bf8.41 & 34.5 & 23.8 & 21.2 \\
\rowcolor{myblue}
\textbf{\method} & \checkmark & & & 
 & \underline{8.78} & \bf46.8 & \bf34.3 & \bf28.2 \\
\bottomrule
\end{tabular}}}
\vspace{-15pt}
\label{tab:rxr_cross}
\end{table}

%% file: tables/scanqa.tex
\begin{table}[t]
    \small
    \centering
    \caption{Evaluation of spatial scene understanding performance on the ScanQA dataset~\citep{azuma2022scanqa} Validation split. \method outperforms current state-of-the-art VLA models and demonstrates superior performance to other 3D LMMs that require additional input, such as depth or camera pose. Note that $*$ indicates 3D LMMs that require task-specific fine-tuning on the ScanQA dataset.}
    % \vspace{-5pt}
    \setlength{\tabcolsep}{1.4pt}
    \scalebox{0.97}{
{\fontsize{8pt}{9pt}\selectfont
\begin{tabular}{lcccccc}
\toprule
& \multicolumn{5}{c}{ScanQA Validation}\\
\cmidrule(lr){2-6}
& Bleu-4 $\uparrow$ & Rouge $\uparrow$ & Cider $\uparrow$ & Meteor $\uparrow$ & EM $\uparrow$ \\
\midrule
\multicolumn{6}{l}{\bf\emph{Task-specific Specialist}}\\
VoteNet+MCAN~\citep{yu2019deep} & 6.2 & 29.8 & 54.7 & 11.4 & 17.3 \\
ScanRefer+MCAN~\citep{yu2019deep} & 7.9 & 30.0 & 55.4 & 11.5 & 18.6 \\
ScanQA~\citep{azuma2022scanqa}  & 10.1 & 33.3 & 64.9 & 13.1 & 21.0 \\
3D-VisTA~\citep{zhu20233d} & 10.4 & 35.7 & 69.6 & 13.9 & 22.4 \\
\midrule
\multicolumn{6}{l}{\bf\emph{3D Large Multi-modal Models}}\\
3D-LLM$_{(flamingo)}$$^{*}$~\citep{hong20233d}  & 7.2 & 32.3 & 59.2 & 12.2 & 20.4 \\
3D-LLM$_{(BLIP2-flant5)}$$^{*}$~\citep{hong20233d}  & 12.0 & 35.7 & 69.4 & 14.5 & 20.5 \\
LL3DA$^{*}$~\citep{chen2024ll3da} & 13.5 & 37.3 & 76.8 & 15.9 & - \\
Chat-3Dv2$^{*}$~\citep{huang2023chat} & 14.0 & - & 87.6 & - & - \\
Scene-LLM$^{*}$~\citep{fu2024scene} & 12.0 & 40.0 & 80.0 & 16.6 & 27.2 \\
LEO~\citep{huang2023embodied} & 13.2 & 49.2 & 101.4 & 20.0 & 24.5 \\
% LLaVA-3D~\citep{zhu2024llava} & 14.5 & 50.1 & 91.7 & 20.7 & 27.0 \\
\midrule
\multicolumn{6}{l}{\bf\emph{2D Vision-Langauge-Action Models}}\\
NaviLLM~\citep{zheng2024towards} & 12.0 & 38.4 & 75.9 & 15.4 & 23.0 \\
% NaVILA$^\dagger$ & 9.6 & 31.9 & 61.3 & 12.8 & 18.4 \\
\rowcolor{myblue}
\textbf{\method} (8 frames) & 14.8 & 46.4 & 95.1 & 18.7 & 27.0 \\
\rowcolor{myblue}
\textbf{\method} (64 frames) & \bf16.9 & \bf49.3 & \bf102.7 & \bf20.1 & \bf28.6 \\
\bottomrule
\end{tabular}}}
\vspace{-15pt}
\label{tab:scanqa}
\end{table}

%% file: tables/r2r_isaac.tex
% \begin{figure*}
%   \begin{minipage}[b]{.36\linewidth}
%     \centering
%     \vspace{1em}
%      \includegraphics[width=1.0\textwidth]{images/misc/vlnce_isaac_vis.pdf}
%       \vspace{-1.5em}
%       \captionof{figure}{VLN-CE-Isaac Benchmark}
%   \label{fig:issac-vis}
%   \end{minipage}
%   \hspace{0.4em}
%       \begin{minipage}[b]{.54\linewidth}
%       % \vspace{-1em}
%         \small
%         \centering
%         \setlength{\tabcolsep}{1.2pt}
%         \captionof{table}{VLN-CE-Isaac evaluation.}
%         \vspace{-0.5em}
%         \scalebox{0.97}{
%     {\fontsize{8pt}{9pt}\selectfont
% \begin{tabular}{lcccccccc}
% \toprule
% & \multicolumn{3}{c}{Low-level Observation} & & \multicolumn{4}{c}{\bench} \\
% \cmidrule(lr){2-4} \cmidrule(lr){6-9}
% & Proprio. & LiDAR & Height Scan & & NE $\downarrow$ & OS $\uparrow$ & SR $\uparrow$ & SPL $\uparrow$ \\
% \midrule
% \rowcolor{mygray}
% Oracle & & &  & & 5.25 & 59.8 & 51.3 & 46.9 \\
% \midrule
% NaVILA-Go2-Blind & \checkmark & & & & 6.03 & 49.0 & 36.2 & 33.3 \\
% \rowcolor{myblue}
% NaVILA-Go2-Vision & \checkmark & \checkmark & & & \textbf{5.49} & \textbf{58.7} & \textbf{50.2} &  \textbf{45.5}\\
% \midrule
% NaVILA-H1-Blind & \checkmark & & & &  7.67 & 33.3 & 24.4 & 21.0\\
% \rowcolor{myblue}
%  NaVILA-H1-Vision & \checkmark & & 
% \checkmark & & \textbf{5.86}& \textbf{54.6} & \textbf{45.3} & \textbf{40.3} \\
% \bottomrule
% \end{tabular}}}
%     \label{tab:r2r_isaac}
%       \end{minipage}
% % \vspace{-10pt}
% \end{figure*}

\begin{figure}[!t]
\centering
% \vspace{1em}
\includegraphics[width=0.36\textwidth]{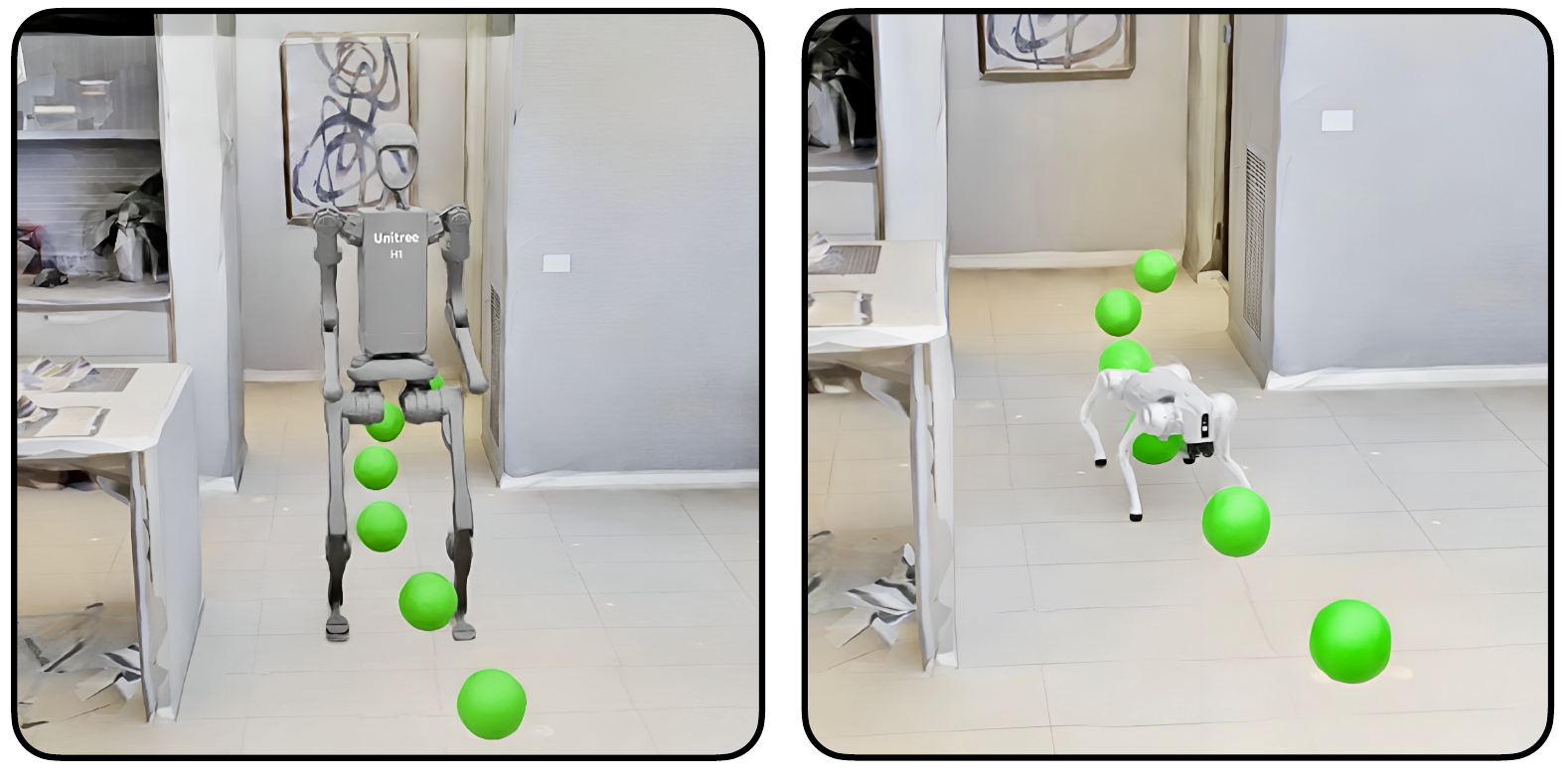}
\vspace{-0.5em}
\caption{VLN-CE-Isaac Benchmark visualization.}
\vspace{-5pt}
\label{fig:issac-vis}
\end{figure}

\begin{table}[!t]
\centering
\small
\setlength{\tabcolsep}{2pt}
\caption{VLN-CE-Isaac evaluation results.}
\vspace{-0.5em}
\scalebox{0.97}{
{\fontsize{8pt}{9pt}\selectfont
\begin{tabular}{lcccccccc}
\toprule
& \multicolumn{3}{c}{Low-level Observation} & & \multicolumn{4}{c}{\bench} \\
\cmidrule(lr){2-4} \cmidrule(lr){6-9}
& Proprio. & LiDAR & Height Scan & & NE $\downarrow$ & OS $\uparrow$ & SR $\uparrow$ & SPL $\uparrow$ \\
\midrule
\rowcolor{mygray}
\textit{Oracle} & & &  & & \textit{5.25} & \textit{59.8} & \textit{51.3} & \textit{46.9} \\
\midrule
\multicolumn{9}{l}{\raisebox{-0.2ex}{\includegraphics[height=1em]{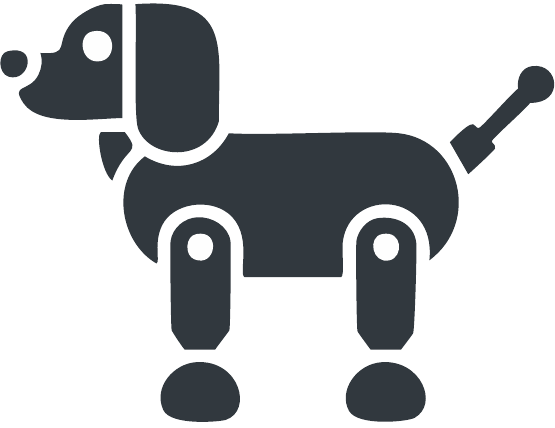}} \bf\emph{Unitree Go2}} \\
NaVILA-Blind & \checkmark & & & & 6.03 & 49.0 & 36.2 & 33.3 \\
\rowcolor{myblue}
\textbf{NaVILA-Vision} & \checkmark & \checkmark & & & \textbf{5.49} & \textbf{58.7} & \textbf{50.2} &  \textbf{45.5}\\
\midrule
\multicolumn{9}{l}{\raisebox{-0.2ex}{\includegraphics[height=1em]{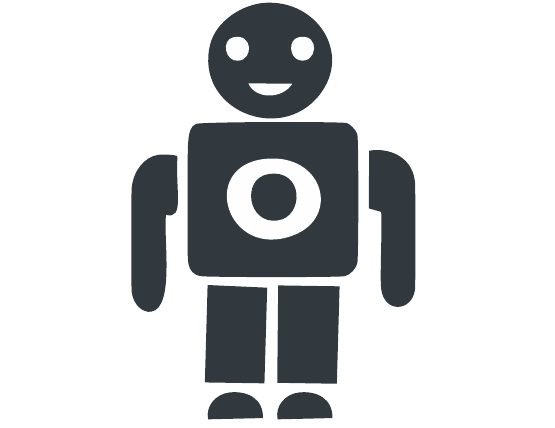}} \bf\emph{Unitree H1}} \\
NaVILA-Blind & \checkmark & & & &  7.67 & 33.3 & 24.4 & 21.0\\
\rowcolor{myblue}
\textbf{NaVILA-Vision} & \checkmark & & \checkmark & & \textbf{5.86}& \textbf{54.6} & \textbf{45.3} & \textbf{40.3} \\
\bottomrule
\end{tabular}}}
\vspace{-5pt}
\label{tab:r2r_isaac}
\end{table}

%% file: tables/low_level.tex
\begin{table}[!t]
\centering
\small
\setlength{\tabcolsep}{9pt}
\caption{Low level policy performance.}
\vspace{-0.5em}
\scalebox{0.97}{
{\fontsize{8pt}{9pt}\selectfont
\begin{tabular}{lccc}
\toprule
 & \makecell{Linear Vel. \\ Error $\downarrow$} & \makecell{Angular Vel. \\ Error $\downarrow$} & \makecell{Collision \\ Rate $\downarrow$} \\
\midrule
ROA$_{(w/ BC Loss)}$~\cite{fu2022deep} & 0.189 & 0.152 & 3.25 \\
ROA~\cite{fu2022deep} & 0.161 & 0.152 & 3.09 \\
\midrule
\rowcolor{myblue}
\textbf{\method} & \textbf{0.066} & \textbf{0.113} & \textbf{0.81} \\
\bottomrule
\end{tabular}}}
\vspace{-15pt}
\label{tab:low_level}
\end{table}

%% file: tables/real_eval.tex
\begin{table}[!t]
\small
\centering
\setlength{\tabcolsep}{1.5pt}
\caption{\label{tab:real_eval}Real-world experiments on quadruped (Unitree Go2) and humanoid (Booster T1) conducted in different environments (\texttt{Workspace}, \texttt{Home}, and \texttt{Outdoor}). Simple and Complex refer to simple and complex instruction-following tasks, respectively. Note that ${\dagger}$ indicates models trained without human touring videos.}
\vspace{-0.5em}
\scalebox{0.97}{
{\fontsize{8pt}{9pt}\selectfont
\begin{tabular}{l l cccc cccc cccc}
\toprule
& & \multicolumn{4}{c}{\texttt{Workspace}} & \multicolumn{4}{c}{\texttt{Home}} & \multicolumn{4}{c}{\texttt{Outdoor}} \\
\cmidrule(lr){3-6} \cmidrule(lr){7-10} \cmidrule(lr){11-14}
& & \multicolumn{2}{c}{Simple} & \multicolumn{2}{c}{Complex} & \multicolumn{2}{c}{Simple} & \multicolumn{2}{c}{Complex} & \multicolumn{2}{c}{Simple} & \multicolumn{2}{c}{Complex} \\
\cmidrule(lr){3-4} \cmidrule(lr){5-6} \cmidrule(lr){7-8} \cmidrule(lr){9-10} \cmidrule(lr){11-12} \cmidrule(lr){13-14}
& & NE$\downarrow$ & SR$\uparrow$ & NE$\downarrow$ & SR$\uparrow$ & NE$\downarrow$ & SR$\uparrow$ & NE$\downarrow$ & SR$\uparrow$ & NE$\downarrow$ & SR$\uparrow$ & NE$\downarrow$ & SR$\uparrow$ \\ 
\midrule
\multicolumn{14}{l}{\raisebox{-0.2ex}{\includegraphics[height=1em]{images/misc/icon_dog.pdf}} \bf\emph{Unitree Go2}} \\
GPT-4o~\cite{openai2024gpt4o} & & 2.01 & \underline{0.67} & 2.38 & 0.33 & \underline{1.49} & \underline{0.53} & 3.00 & 0.00 & - & \underline{0.67} & - & \underline{0.50} \\
\method$\!^{\dagger}$ & & \underline{2.00} & 0.60 & \underline{1.81} & \underline{0.73} & 2.17 & 0.47 & \underline{2.32} & \underline{0.40} & - & 0.00 & - & 0.00 \\
\rowcolor{myblue}
\textbf{\method} & & \textbf{1.29} & \textbf{1.00} & \textbf{1.76} & \textbf{0.80} & \textbf{1.15} & \textbf{1.00} & \textbf{1.76} & \textbf{0.67} & - & \textbf{1.00} & - & \textbf{0.83} \\ 
\midrule
\multicolumn{14}{l}{\raisebox{-0.2ex}{\includegraphics[height=1.em]{images/misc/icon_humanoid.pdf}} \bf\emph{Booster T1}} \\
GPT-4o~\cite{openai2024gpt4o} & & \underline{1.53} & \underline{0.67} & 2.78 & 0.13 & - & - & - & - & - & \underline{0.44} & - & - \\
\method$\!^{\dagger}$ & & 2.36 & 0.40 & \underline{2.16} & \underline{0.62} & - & - & - & - &  - & 0.22 & - & - \\
\rowcolor{myblue}
\textbf{\method} & & \textbf{1.18} & \textbf{0.93} & \textbf{1.91} & \textbf{0.67} & - & - & - & - & - & \textbf{0.89} & - & - \\ 
\bottomrule
\end{tabular}
}
}
\vspace{-15pt}
\end{table}

%% file: sections/02-Related-Work.tex
\section{Related Work}
\label{related_work}
\myparagraph{Visual Navigation.}
Visual navigation has been a long-standing research topic in robotics~\citep{moravec1980obstacle, elfes1987sonar, thrun2001robust, gervet2023navigating}. Classical methods rely on pre-computed~\citep{thrun1999minerva} or geometric maps built with depth sensors~\citep{newcombe2011kinectfusion} or monocular cameras while localizing the robot (SLAM)~\citep{davison2007monoslam, jones2011visual}. More recently, learning-based approaches using Imitation Learning~\citep{chaplot2018gated, codevilla2018end} and Reinforcement Learning~\citep{mnih2015human, lillicrap2015continuous} have demonstrated strong performance and expanded applications to Vision-Language Navigation.
% Visual navigation has been a long-standing research topic in robotics for decades~\citep{moravec1980obstacle,elfes1987sonar,thrun2001robust,gervet2023navigating}. Classical approaches often rely on pre-computed maps~\citep{thrun1999minerva} or build geometric maps of the environment using depth sensors~\citep{newcombe2011kinectfusion} or monocular RGB cameras while localizing the robot simultaneously (SLAM)~\citep{davison2007monoslam,jones2011visual}. Recently, learning-based approaches with Imitation Learning~\citep{chaplot2018gated,codevilla2018end} and Reinforcement Learning~\citep{mnih2015human,lillicrap2015continuous} have not only shown impressive results but also enabled wider applications including Vision-Language Navigation.

\myparagraph{Vision-Language Navigation.}
Vision-Language Navigation (VLN) is a fundamental challenge in embodied AI, where agents navigate complex environments using visual cues and natural language instructions. The field has evolved significantly over time. Early research~\citep{anderson2018vision,rxr,qi2020reverie} focused on discrete navigation in simulated environments like MP3D~\citep{Matterport3D}, where agents teleport between predefined nodes on a navigation graph~\citep{tan2019learning,fried2018speaker,ma2019selfmonitoring,ke2019tactical,hong2020language,chen2021history,chen2024mapgpt,zhou2024navgpt}. As foundation models advanced, many VLN systems improved dramatically by leveraging large-scale pre-trained models~\citep{majumdar2020improving, li2019robust} and pre-training techniques~\citep{guhur2021airbert,wang2023scaling,kamath2023new}, approaching human-level performance in this setting. However, this setup emphasized high-level decision-making while neglecting the challenges of underlying motion control. Recently, research~\citep{zhang2024navid, raychaudhuri2021language,georgakis2022cross, chen2022weakly,chen2024affordances} has shifted towards continuous environments (VLN-CE~\citep{krantz2020beyond}) using simulators like Habitat~\citep{savva2019habitat}. This introduces greater complexity, as agents must perform mid-level actions such as moving forward or rotating, rather than teleporting between nodes. To bridge the gap between discrete and continuous navigation, some approaches~\citep{,krantz2022sim,an2024etpnav, an2023bevbert, irshad2021hierarchical} use simulator pre-trained waypoint models~\citep{hong2022bridging,krantz2021waypoint} that predict candidate positions around the agent and have shown significant performance gains. However, they often struggle to generalize due to their reliance on simulator-specific data. Additionally, the candidate positions predicted by these models only cover nearby locations and do not account for low-level motion planning or obstacle avoidance. In this paper, we aim to advance VLN toward real-world robotics applications, particularly for challenging-legged robots. \method handles both high-level decision-making and generates low-level actions to control the robot's full motion. Additionally, we introduce a new VLN benchmark built on Isaac Sim, offering a more realistic simulation environment, which we believe will benefit future work in VLN.

\myparagraph{Robot Foundation Models.} Robot foundation models aim to provide a unified framework that processes inputs from various modalities (e.g., vision and language) and directly outputs actions to enable robots to perform complex tasks. Existing works \citep{brohan2023rt, kim2024openvla, team2024octo} trained on large-scale robotic datasets to get general robot policies, but mainly focusing on manipulation tasks. \citet{doshiscaling} and \citet{yang2024pushing} proposed end-to-end visual-language cross-embodiment models for different robotic tasks. Recently, several foundational navigation models have been proposed~\citep{zeng2024poliformer,shah2023gnm,sridhar2024nomad}. However, they mainly focus on goal navigation with either short language descriptions or a target image as input. As for legged robots, \citet{ding2023quar} proposed a unified model to leverage vision and language inputs and generate executable low-level actions. Another line of work~\citep{chen2024commonsense,ouyang2024long} focuses on training specialized policies as a skill bank to handle specific actions, with either a VLM or LLM serving as the controller to decide which skill to execute. Similarly, these methods cannot perform instruction-following tasks, as they struggle to comprehend complex instructions that are essential for general navigation. To address this, we propose a VLA model specifically designed for general vision language navigation tasks. 

% Our model generates high-level action commands, which are then executed by a low-level policy. This approach enables the robot to interpret complex instructions and navigate effectively towards the goals.

\myparagraph{Legged Robot Locomotion Learning.}
Legged robot locomotion learning focuses on enabling robots to traverse various terrains. Previous works \citep{wang2023learning, long2024hybrid} rely solely on robot's proprioceptive information struggle in scenarios like obstacle avoidance. Other end-to-end vision-based approaches \citep{kareer2023vinl, yang2021learning, imai2022vision, yang2023neural} are vulnerable to extreme environmental conditions, such as intense sunlight, due to the limitations of sensors. \citet{lee2020learning} incorporate LiDAR sensors in addition to depth cameras to improve terrain sensing, but rely on time-inefficient two-stage training. Additionally, during training, \citet{miki2022learning} query predefined terrain heights to construct a height map and then rely on an external tool\cite{miki2022elevation} for height map generation during deployment, resulting in discrepancies between training and deployment. To overcome these limitations, we propose a single-stage RL framework that integrates LiDAR sensing inputs during training, allowing the robot to directly learn from interacting with the environments for better efficiency and robustness in complex scenarios. 

%% file: sections/05-Conclusion.tex
\vspace{-0.2em}
\section{Conclusion and Limitations}
\vspace{-0.2em}
We introduce \method, a two-level framework that unifies VLAs with locomotion skills for generic navigation. \method generates high-level language commands while a real-time locomotion policy handles obstacle avoidance, enhancing robustness across robots. This design preserves reasoning, prevents overfitting, and enables direct learning from human videos for better generalization. \method achieves a $17\%$ gain on classic VLN benchmarks, outperforms distillation-based low-level policies, surpasses vision-blind policies on VLN-CE-Isaac1K, and demonstrates strong real-world performance across diverse environments and legged robots.

% \clearpage
\myparagraph{Limitations.}
While \method demonstrates strong performance, it fails in some real-world cases (see Appx. Sec.~\ref{sup:limitations}). Enhancing generalizability and spatial understanding through larger-scale training in realistic simulations could help. Additionally, image-based vision-language models are computationally intensive. Advances in long-context LLMs may alleviate this by enabling more efficient sequence processing.

% We introduce \method, a powerful two-level framework that unifies vision-language-action models (VLAs) with locomotion skills for generic navigation tasks. \method generates high-level, language-based commands, while a real-time locomotion policy handles obstacle avoidance. This dual-frequency design improves robustness and flexibility across different robots. By preserving reasoning capabilities through language-based actions, \method avoids overfitting and can be trained on broader tasks. Moreover, our two-level design enables VLAs to learn from human videos, enhancing generalizability by leveraging diverse real-world demonstrations. In experiments, \method shows a $17\%$ improvement on classic VLN benchmarks, outperforms vision-blind policies in our new VLN-CE-Isaac1K benchmark, and demonstrates strong real-world performance across diverse scenes.

%% file: sections/06-Appendix.tex
\clearpage
\appendix

\setcounter{page}{1} % Reset page counter

% Enable ToC
\hypersetup{linkcolor=citecolor}

\cftsetindents{subsection}{0.em}{1.em} % Adjust the first value to reduce indent
\cftsetindents{subsubsection}{1em}{2em} % Adjust the first value to reduce indent

\section*{Table of Contents} 
\startcontents[sections]
\printcontents[sections]{l}{1}{\setcounter{tocdepth}{3}}

\hypersetup{
    urlcolor=blue, colorlinks,citecolor=citecolor, linkcolor=red
}
% % rest to original KM blue
% \newpage

% \twocolumn
\subsection{More Ablation Studies}

\subsubsection{Different Simulation Data Blends}

We perform an ablation study to assess the impact of different simulation data blends on training VLA. As shown in Table~\ref{tab:abl}, training navigation data without label rebalancing leads to a significant drop in performance. Additionally, training VLA exclusively on RxR data demonstrates reasonable cross-dataset performance on R2R-CE, supporting our observations in Table~\ref{tab:rxr_cross}. Lastly, we investigate whether excluding RxR dataset degrades R2R-CE performance. The results suggest that the RxR dataset does not significantly contribute to the R2R-CE performance.

\begin{table}[!ht]
    \small
    \centering
\vspace{-0.5em}
    \caption{Results on R2R-CE using different data blends.}
\vspace{-0.5em}
    \setlength{\tabcolsep}{4.4pt}
    \scalebox{0.97}{
{\fontsize{8pt}{9pt}\selectfont
\begin{tabular}{lccccc}
\toprule
& \multicolumn{4}{c}{R2R-CE Val Unseen}\\
\cmidrule(lr){2-5}
& NE $\downarrow$ &  OSR $\uparrow$ & SR $\uparrow$ & SPL $\uparrow$ \\
\midrule
\method$\!^{\dagger}$ (w/o Label Balancing) & 7.82 & 47.5 & 30.0  & 25.1 \\
\method$\!^{\dagger}$ (w/ RxR only) & 7.57 & 40.8 & 31.5  & 27.8 \\
\method$\!^{\dagger}$ (w/o RxR)  & 6.11 & 57.0 & 47.7  & 42.4 \\
\rowcolor{myblue}
\method$\!^{\dagger}$ & \bf5.37 & \bf57.6 & \bf49.7 & \bf45.5 \\
\bottomrule
\end{tabular}}}
\vspace{-0.1em}
\label{tab:abl}
\end{table}

\subsubsection{Human Touring Video Data}
\label{abl:real-sim}
We perform ablation studies in simulations to assess the effect of using real-world data from YouTube human touring videos. As shown in Table~\ref{tab:abl:real:r2r}, incorporating this data leads to significant performance gains, with approximately 5\% improvements in OS, SR, and SPL metrics. Similarly, results from real-world experiments (Table~\ref{tab:real_eval}) demonstrate higher success rates and fewer navigation errors. These results validate the effectiveness of our data pipeline and highlight the scalability of NaVILA’s framework, which supports easy integration of data from diverse sources.

\begin{table}[!ht]
    \small
    \centering
\vspace{-0.5em}
    \caption{Results on R2R-CE using additional real data from human touring videos.}
\vspace{-0.5em}
    \setlength{\tabcolsep}{4.4pt}
    \scalebox{0.97}{
{\fontsize{8pt}{9pt}\selectfont
\begin{tabular}{lccccc}
\toprule
& \multicolumn{4}{c}{R2R-CE Val Unseen}\\
\cmidrule(lr){2-5}
& NE $\downarrow$ &  OSR $\uparrow$ & SR $\uparrow$ & SPL $\uparrow$ \\
\midrule
\method$\!^{\dagger}$ & 5.37 & 57.6 & 49.7 & 45.5 \\
\rowcolor{myblue}
\method & \bf5.22 & \bf62.5 & \bf54.0 & \bf49.0 \\
\bottomrule
\end{tabular}}}
\vspace{-0.5em}
\label{tab:abl:real:r2r}
\end{table}

\blfootnote{\footnotesize ${\dagger}$ indicates models trained without human touring videos.}

% \begin{table}[!ht]
% \small
% \centering
% \setlength{\tabcolsep}{5.pt}
% \vspace{-0.5em}
% \caption{\label{tab:able:real:real}Ablation study on real video data in real-world experiments. Simple and Complex refer to simple and complex instruction-following tasks, respectively.}
% \vspace{-0.8em}
% \scalebox{0.97}{
% {\fontsize{8pt}{9pt}\selectfont
% \begin{tabular}{l l cccc cccc cccc}
% \toprule
% & & \multicolumn{4}{c}{\texttt{Laboratory}} & \multicolumn{4}{c}{\texttt{House}} & \multicolumn{4}{c}{\texttt{Outdoor}} \\
% \cmidrule(lr){3-6} \cmidrule(lr){7-10} \cmidrule(lr){11-14}
% & & \multicolumn{2}{c}{Simple} & \multicolumn{2}{c}{Complex} & \multicolumn{2}{c}{Simple} & \multicolumn{2}{c}{Complex} & \multicolumn{2}{c}{Simple} & \multicolumn{2}{c}{Complex} \\
% \cmidrule(lr){3-4} \cmidrule(lr){5-6} \cmidrule(lr){7-8} \cmidrule(lr){9-10} \cmidrule(lr){11-12} \cmidrule(lr){13-14}
% & & NE$\downarrow$ & SR$\uparrow$ & NE$\downarrow$ & SR$\uparrow$ & NE$\downarrow$ & SR$\uparrow$ & NE$\downarrow$ & SR$\uparrow$ & NE$\downarrow$ & SR$\uparrow$ & NE$\downarrow$ & SR$\uparrow$ \\ 
% \midrule
% \method(w/o Real Videos) & & 2.00 & 0.60 & 1.81 & 0.73 & 2.17 & 0.47 & 2.32 & 0.40 & - & 0.00 & - & 0.00 \\
% \rowcolor{myblue}
% \method & & \textbf{1.29} & \textbf{1.00} & \textbf{1.76} & \textbf{0.80} & \textbf{1.15} & \textbf{1.00} & \textbf{1.76} & \textbf{0.67} & \textbf{-} & \textbf{1.00} & \textbf{-} & \textbf{0.83} \\ 
% \bottomrule
% \end{tabular}
% }
% }
% \vspace{-0.5em}
% \end{table}

\subsubsection{Different Memory Sizes}
We conduct an ablation study to evaluate the impact of memory size (number of history frames) on the navigation task using R2R-CE benchmark. The results in Table~\ref{tab:abl:frame:r2r} show that for R2R-CE, 8 frames are sufficient to cover most instruction horizons, with limited performance gains from increasing the memory size. For real-world experiments, we use an 8-frame memory size due to latency constraints. 

\begin{table}[!ht]
    \small
    \centering
    \vspace{-0.5em}
    \caption{Ablation study on different memory size using R2R-CE~\citep{r2r} Validation Unseen split.}
    \vspace{-0.5em}
    \setlength{\tabcolsep}{8.4pt}
    \scalebox{0.97}{
{\fontsize{8pt}{9pt}\selectfont
\begin{tabular}{lccccc}
\toprule
& \multicolumn{4}{c}{R2R-CE Val Unseen}\\
\cmidrule(lr){2-5}
& NE $\downarrow$ &  OSR $\uparrow$ & SR $\uparrow$ & SPL $\uparrow$ \\
\midrule
NaVid~\citep{zhang2024navid} & 5.47 & 49.0 & 37.0 & 35.0 \\
\midrule
\rowcolor{myblue}
\method$\!^{\dagger}$ (8 frames) &  \bf5.37 & \underline{57.6} & \underline{49.7} & \bf45.5 \\
\method$\!^{\dagger}$ (16 frames) & 5.63 & 55.8 & 48.6 & 44.4 \\
\method$\!^{\dagger}$ (32 frames) & 5.74 & 55.9 & 49.5 & 44.1 \\
\method$\!^{\dagger}$ (64 frames) & 5.63 & \bf60.5 & \bf50.1 & \underline{45.4} \\
\bottomrule
\end{tabular}}}
\vspace{-1.5em}
\label{tab:abl:frame:r2r}
\end{table}

% \begin{table}[!ht]
%     \small
%     \centering
%     \caption{Ablation study on different memory size using ScanQA dataset~\citep{azuma2022scanqa} Validation split.}
%     % \vspace{-0.8em}
%     \setlength{\tabcolsep}{4.4pt}
%     \scalebox{0.97}{
% {\fontsize{8pt}{9pt}\selectfont
% \begin{tabular}{lcccccc}
% \toprule
% & \multicolumn{5}{c}{ScanQA Validation}\\
% \cmidrule(lr){2-6}
% & Bleu-4 $\uparrow$ & Rouge $\uparrow$ & Cider $\uparrow$ & Meteor $\uparrow$ & EM $\uparrow$ \\
% \midrule
% \multicolumn{6}{l}{\bf\emph{3D Large Multi-modal Models}}\\
% Scene-LLM~\citep{fu2024scene} & 12.0 & 40.0 & 80.0 & 16.6 & 27.2 \\
% LEO~\citep{huang2023embodied} & 13.2 & 49.2 & 101.4 & 20.0 & 24.5 \\
% \midrule
% \multicolumn{6}{l}{\bf\emph{2D Vision-Langauge-Action Models}}\\
% NaviLLM~\citep{zheng2024towards} & 12.0 & 38.4 & 75.9 & 15.4 & 23.0 \\
% \method(8 frames) & 14.8 & 46.4 & 95.1 & 18.7 & 27.0 \\
% \method(16 frames) & 15.2 & 48.3 & 99.8 & 19.6 & 27.4 \\
% \method(32 frames) & 16.1 & 49.4 & 101.6 & 20.2 & 28.1 \\
% \rowcolor{myblue}
% \method(64 frames) & \bf16.9 & \bf49.3 & \bf102.7 & \bf20.1 & \bf28.6 \\
% \bottomrule
% \end{tabular}}}
% \label{tab:abl:frame:scanqa}
% \end{table}

\subsection{More Qualitative Results}
\subsubsection{\bench}
\label{appendix_vln_ce_isaac}
\begin{figure}[!ht]
    \centering
    \vspace{-0.5em}
    \includegraphics[width=0.85\linewidth]{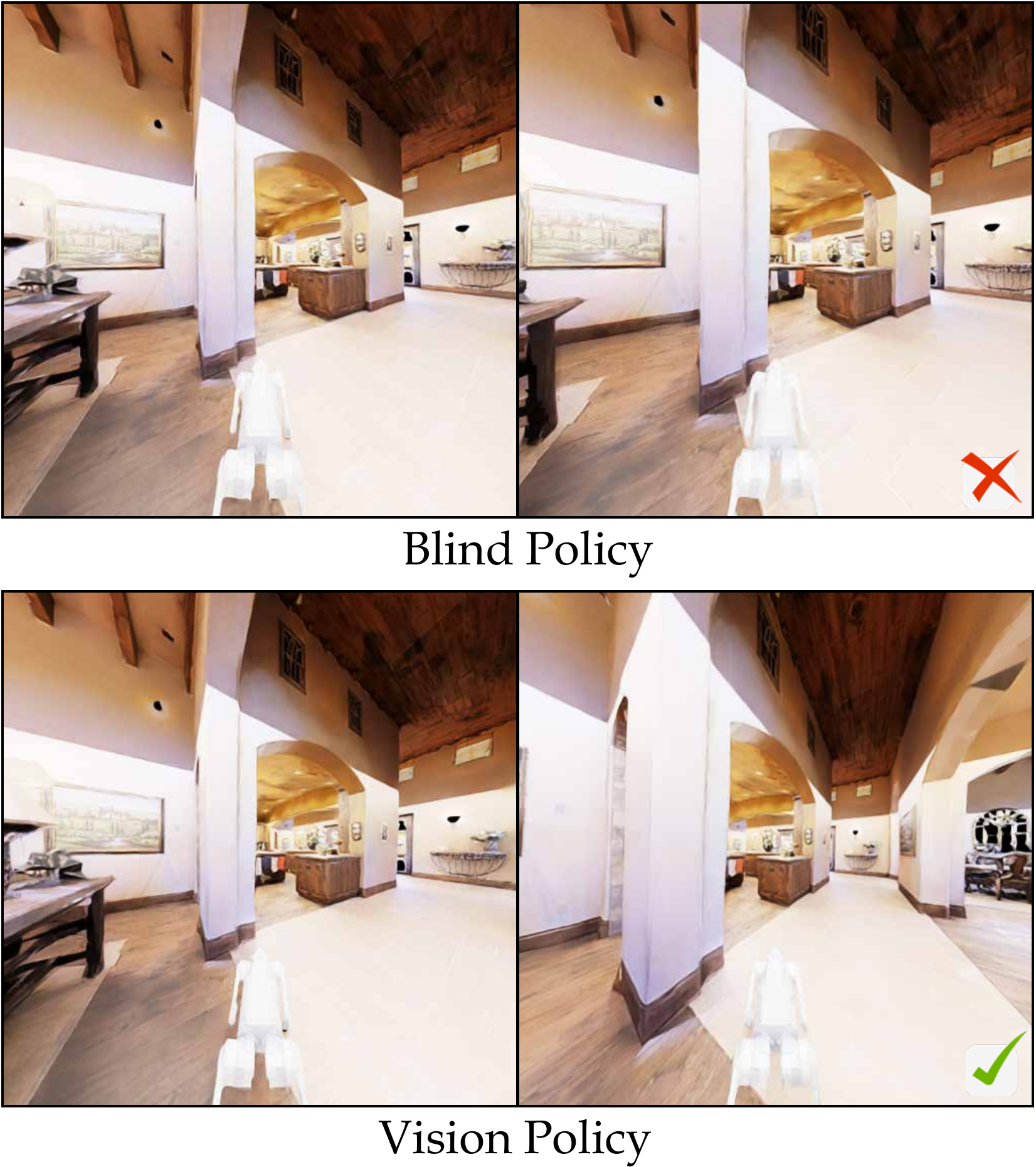}
    \vspace{-0.5em}
    \caption{Comparison between Go2 blind policy and vision policy. The blind policy failed to avoid the obstacles and got stuck. The vision policy detected the obstacle and got around to avoiding it.}
    \vspace{-0.5em}
    \label{fig:vlnce_isaac_appendix}
\end{figure}

Here we show a visualization example highlighting why the Go2 vision policy significantly outperforms the blind policy. As demonstrated in Figure~\ref{fig:vlnce_isaac_appendix}, when encountering an obstacle, the VLA, which is not specifically trained for obstacle avoidance, failed to navigate around it effectively. The blind policy, following the VLA's commands without additional sensory input, became stuck at the obstacle. In contrast, the vision-based policy, trained to handle obstacles using LiDAR input, can autonomously avoid dangers even when the high-level VLA model does not detect them.
% Images showing the obstacle avoidance capability by 

\subsubsection{Real-World}
\begin{figure*}[!ht]
\vspace{-0.5em}
    \centering
    \includegraphics[width=0.99\linewidth]{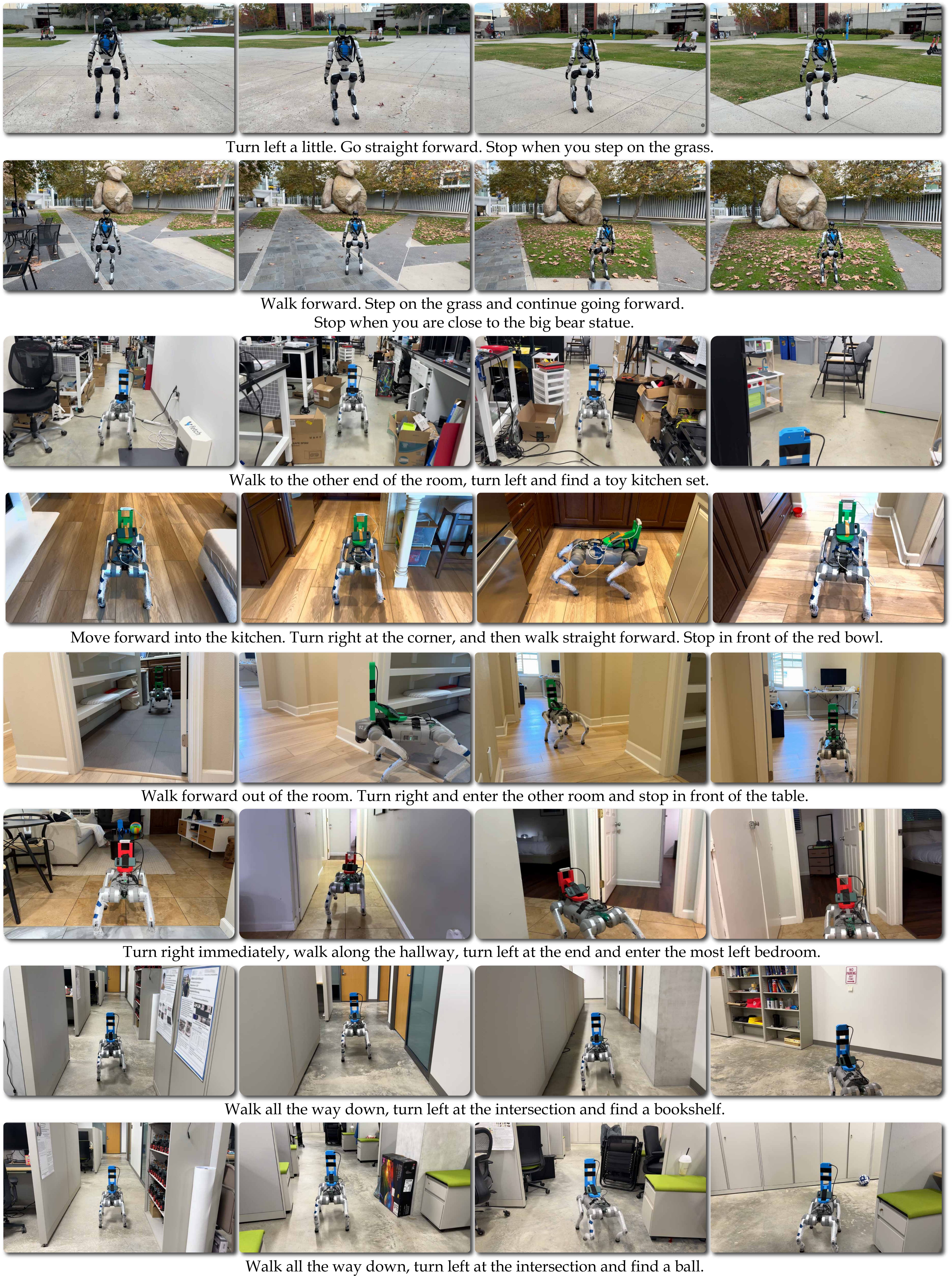}
    \caption{We show more results in diverse environments, such as urban streets, campus sidewalks, courtyards, and different houses. These settings add significant variety and challenges, including challenging terrains, dynamic objects, and different lighting conditions. The results \method achieved represent a significant milestone, showcasing capabilities that have never been demonstrated before.}
    \label{fig:qual_appendix}
\vspace{-1.5em}
\end{figure*}

We show more real-world results in Figure~\ref{fig:qual_appendix}. \method demonstrates robust performance and exceptional generalization across diverse settings.

\clearpage
\onecolumn
\clearpage

% \subsection{More Results on ScanQA}

\twocolumn

\subsubsection{Spatial Scene Understanding}
In Figure~\ref{fig:scanqa_appendix}, we show NaVILA's qualitative results on spatial scene understanding using the ScanQA benchmark. Given a sequence of images sampled from a video, \method can ground and locate objects correctly. 

\begin{center}
    \centering 
    \includegraphics[width=\linewidth]{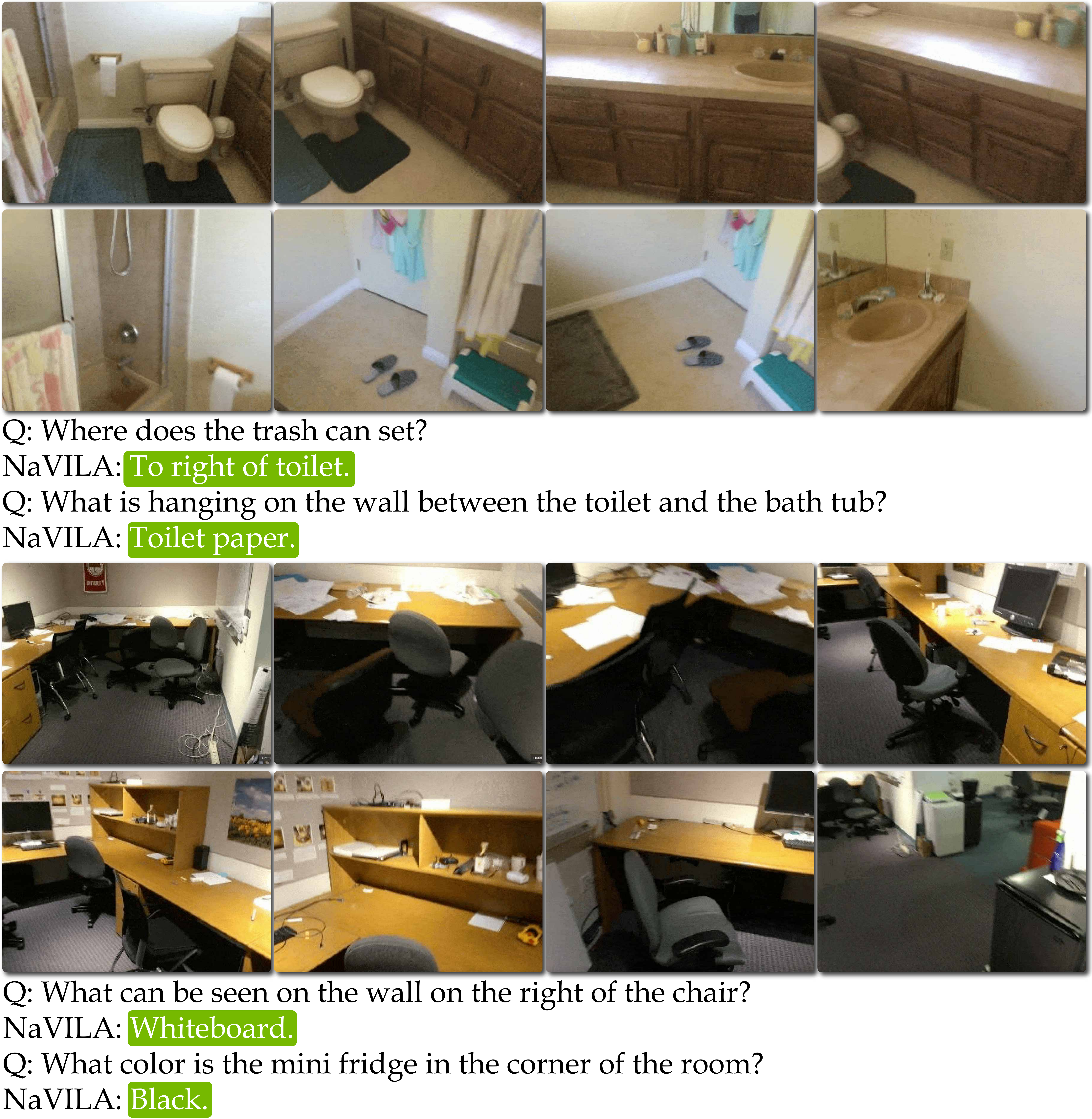}
    \vspace{-1em}
    \captionof{figure}{Spatial scene understanding results on ScanQA benchmark.}
    \label{fig:scanqa_appendix}
\end{center}

\subsection{More Implementation Details}
\subsubsection{Video Navigation Trajectory Summarization}
\label{sup:implementation-details:nav-sum}
We provide the data prompts for our auxiliary task of video navigation trajectory summarization. Following the approach in~\citep{zhang2024navid}, we construct prompt templates that characterize the LLM as a robot designed for navigation. We process the trajectory videos into history frames, insert the frame tokens into the prompt, and ask the LLM to infer the navigation instructions from the video. This task is designed to enhance the robot's scene understanding and its familiarity with the instruction format.

\begin{lstlisting}[
    basicstyle=\ttfamily\footnotesize, 
    backgroundcolor=\color{mygray}, 
    keywordstyle=\textbf, 
    caption={}, 
    label={lst:clip},
    rulecolor=\color{black},
    xleftmargin=\dimexpr\fboxsep-\fboxrule,
]
Assume you are a robot designed for navigation. You 
are provided with captured image sequences:

<frame3><frame6><frame9>...

Based on this image sequence, please describe the 
navigation trajectory of the robot.
\end{lstlisting}

% \begin{table}[!ht]\centering

% \begin{minipage}{0.99\columnwidth}\vspace{0mm}    \centering
%     \centering
%     \small
%      \hspace{-6mm}
%     \begin{tabular}{p{0.99\columnwidth}}

% \begin{minipage}{0.99\columnwidth}\vspace{0mm}
% \texttt{Assume you are a robot designed for navigation. You are provided with captured image sequences: $\langle$frame3$\rangle$$\langle$frame6$\rangle$$\langle$frame9$\rangle$
% Based on this image sequence, please describe the navigation trajectory of the robot.}
% \end{minipage}
%     \end{tabular}
% \end{minipage}
% \end{table}

\subsubsection{VLA Hyperparameters}
\label{sup:vlm-details:hyperparams}
Please refer to VILA's paper for details on the hyperparameters used in the first two stages. In the instruction fine-tuning stage, we use a learning rate of $1e^{-4}$ with cosine decay and a warm-up ratio of $0.03$. We will release both our training code and data upon paper publication.

\subsubsection{Locomotion Motion Policy}
\label{sup:policy-details}

\begin{figure}[!ht]
  \centering
    \centering
    \includegraphics[width=.95\linewidth]{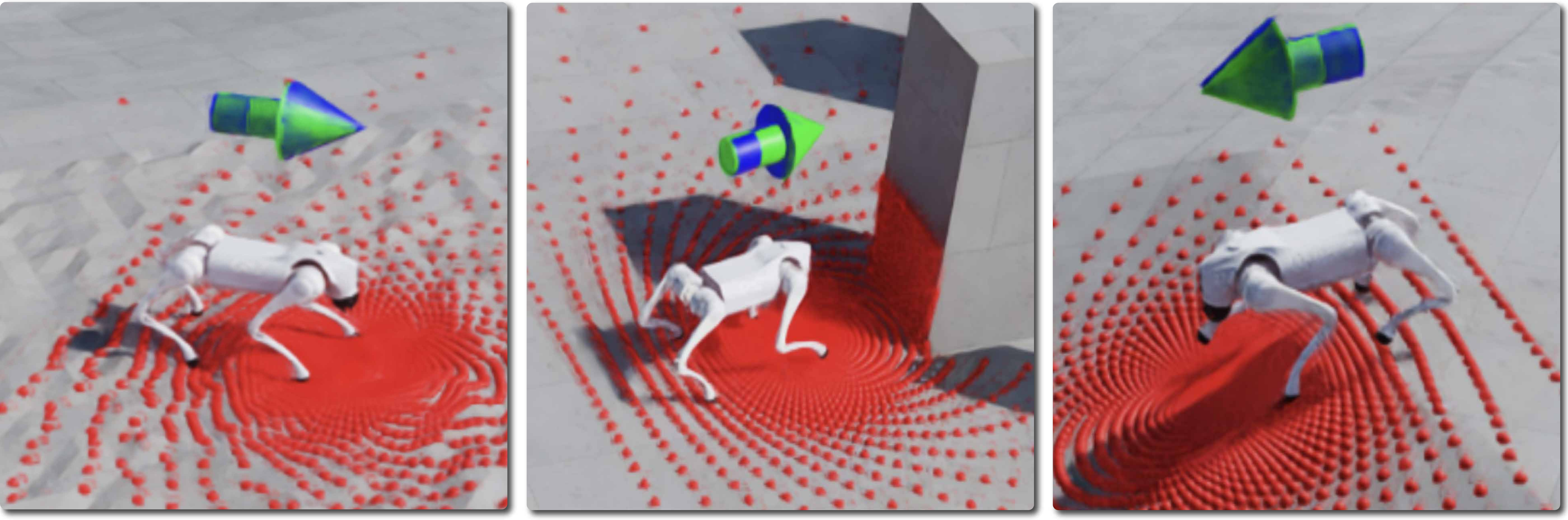}
    \vspace{-0.5em}
    \caption{Random rough, obstacle and slope terrain.}
    \vspace{-0.5em}
    \label{fig:obstacle}
\end{figure}

The reward functions and domain randomization used during Go2 locomotion policy training are listed in Table~\ref{tab:reward} and Table~\ref{tab:domain_rand}.
The robust policy is trained on flat, rough, slope and obstacle terrains shown in Figure~\ref{fig:obstacle}. LiDAR and height map settings are detailed in Table~\ref{tab:lidar_heightmap_settings}.  

\begin{table}[!ht]
    \centering
    \small
    \vspace{-0.5em}
    \caption{Reward function parameters for training RL policy.}
    \vspace{-0.5em}
    \setlength{\tabcolsep}{3.6pt}
    \scalebox{0.97}{
{\fontsize{8pt}{9pt}\selectfont
{
    \begin{tabular}{lll}
    \toprule {Reward} & {Expression}  & {Weight} \\
    \midrule Linear velocity tracking & $\exp(- \|v_{x y}^{\text {cmd }}-v_{x y}\|_2^2  )$ & 1.5 \\
    Angular velocity tracking & $\exp (- \left(\omega_{\text {yaw }}^{\text {cmd }}-\omega_{\text {yaw }}\right)^2)$ & 1.5 \\
    Linear velocity penalty $(z)$ & $v_z^2$ & -2.0 \\
    Angular velocity penalty $(x y)$ & $\lVert\boldsymbol{\omega}_{x y}\|_2^2$ & -0.05 \\
    Flat orientation & $\|\mathbf{g}\|_2^2$ & -2.0 \\
    Joint accelerations & $\|\ddot{\boldsymbol{\theta}}\|^2$ & $-2.5 \times 10^{-7}$ \\
    Energy & $-\lVert\tau \dot{q}\rVert_2^2$ & $-2 \times 10^{-5}$\\
    % \dot{\boldsymbol{\theta}}|^{T}$  \\
    Body height & $\left(h^{\text {target}}-h\right)^2$ & -5.0 \\
    Feet slipping & $-\lVert v_{\text{feet}} \cdot \mathbf{1}[F_{\text{feet}} > 1]\rVert_2$ & 0.05\\
    % Action rate & $\|\mathbf{a}_t-\mathbf{a}_{t-1}\|_2^2$ & -0.02 \\
    % Smoothness & $\|\mathbf{a}_t-2 \mathbf{a}_{t-1}+\mathbf{a}_{t-2}\|_2^2$ & -0.01 \\
    \bottomrule
    \end{tabular}}}}
    \vspace{-0.2em}
    % \caption{Rewards}
    \label{tab:reward}
\end{table}

\begin{table}[!ht]
  \centering
  \small
  \vspace{-0.5em}
  \caption{Domain randomization parameters for training RL policy.}
  \vspace{-0.5em}
    \setlength{\tabcolsep}{10.6pt}
    \scalebox{0.97}{
{\fontsize{8pt}{9pt}\selectfont
{
  \begin{tabular}{ll}
    \toprule
    {Parameter} & {Value} \\ \midrule
   Body Mass & [-3.0, 3.0] \\
    Static Ground Friction & [0.4, 4.0]\\
    Dynamic Ground Friction & [0.4, 4.0] \\
    Motor Strength & [0.9,1.1] \\
    System Delay & [$\Delta_t, \Delta_t$] \\
    \bottomrule
  \end{tabular}}}}
  \vspace{-0.2em}
  \label{tab:domain_rand}
\end{table}

% \textbf{LiDAR and height map specs:}
% \begin{table}[!ht]
% \small
%   \centering
%   \caption{Simulation LiDAR and Height Map Settings}
%   \begin{tabular}{ll}
%     \toprule
%     \textbf{Parameter} & \textbf{Value} \\ \midrule
%     Stiffness & 40 \\
%     Dampness & 0.2\\
%     Horizontal Range (degrees) & (-180, 180) \\
%     Horizontal Resolution (degrees) & 4 \\
%     Voxel Size (m) & 0.06 \\
%     X Range (m) & [-0.8, 0.2] \\
%     Y Range (m) & [-0.8, 0.8] \\
%     Z Range (m) & [0.05, 0.5] \\ \bottomrule
%   \end{tabular}
%   \label{tab:training_specs}
% \end{table}

\begin{table}[!ht]
\small
  \centering
  \vspace{-0.5em}
  \caption{LiDAR and Height Map parameters in simulation.}
  \vspace{-0.5em}
    \setlength{\tabcolsep}{10.6pt}
    \scalebox{0.97}{
{\fontsize{8pt}{9pt}\selectfont
{
  \begin{tabular}{ll}
    \toprule
    {Parameter} & {Value} \\ \midrule
    Channels & 32 \\
    Vertical Range (degrees) & (0, 90) \\
    Horizontal Range (degrees) & (-180, 180) \\
    Horizontal Resolution (degrees) & 4 \\
    Voxel Size (m) & 0.06 \\
    X Range (m) & [-0.8, 0.2] \\
    Y Range (m) & [-0.8, 0.8] \\
    Z Range (m) & [0.05, 0.5] \\ \bottomrule
  \end{tabular}}}}
    \vspace{-1.5em}
  \label{tab:lidar_heightmap_settings}
\end{table}

% \textbf{Obstacle avoidance:}
% Caption for the first type of screenshots

% \subsection{Implementation Details for VLN-CE-Isaac1K}
% \label{sup:implementation-details:benchmark}
% \newpage

\subsubsection{VLA Compute Resources}
\label{sup:compute}
The first two stages of \method are inherited from VILA~\citep{lin2024vila}, which is trained on 16 A100 GPU nodes, with each node having 8 GPUs. The training times for each stage of our 8B model are as follows: connector initialization takes 4 hours, visual language pre-training takes 30 hours. The final visual instruction-tuning stage is experimented on 4 A100 GPU nodes, taking 18 hours. During inference time, the VLA model in \method can be served using a single RTX 4090 GPU with roughly 1 FPS.

\subsection{Parameter-efficient Quantization}
% NaVILA's current wait time between each action is about 1 second, which is practical for real-world deployment. The wait time arises from two factors: image transmission time from Go2 to the server, and the VLA inference time. The transmission time largely depends on the network conditions, while the VLA inference time is approximately 0.6 seconds per sample. 

We explore optimization techniques to improve NaVILA's inference efficiency. Specifically, we apply AWQ~\citep{lin2023awq}, a state-of-the-art quantization method for VLMs, to the FP16 NaVILA-8B model. By converting it to the W4A16 format (low-bit weight-only quantization), we achieved significant improvements: memory requirements dropped by half, and processing speed improved by about 40\%. Most importantly, navigation capabilities remained robust. Results are detailed in Table~\ref{tab:quantization}. These optimizations make NaVILA deployable directly on the robot, which will significantly eliminate image transmission time. We leave this as future work.

\onecolumn

\begin{figure*}[h]
  \centering
  \includegraphics[width=0.99\textwidth]{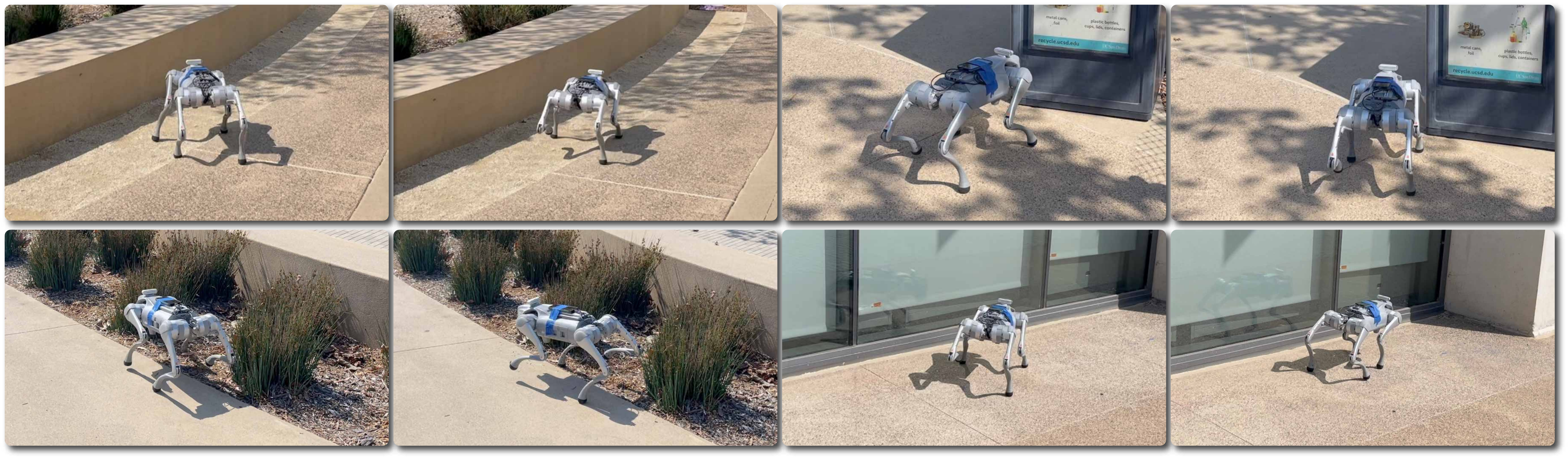}
  \caption{Obstacle avoidance screenshots. Locomotion policy can ensure collision-free in the face of high grass, certain transparent glass, and large objects under strong sunlight. The policy presents robustness on sand and grass terrains.}
  \label{fig:type1}
\end{figure*}

\begin{table*}[h]
    \small
    \centering
    \caption{NaVILA quantization results. The computational cost is tested on RTX 4090 with 1737 context tokens and 10 generated tokens, using a sample from R2R-CE as the test case.}
    % \vspace{-0.5em}
    \setlength{\tabcolsep}{3.6pt}
    \scalebox{0.97}{
{\fontsize{8pt}{9pt}\selectfont
{
\begin{tabular}{lcclcccc}
\toprule
& \multicolumn{2}{c}{Computational Cost} & & \multicolumn{4}{c}{R2R Val-Unseen}\\
\cmidrule(lr){2-3} \cmidrule(lr){5-8}
& Total Latency (ms) $\downarrow$ & GPU Memory (GB) $\downarrow$ & & NE $\downarrow$ & OS $\uparrow$ & SR $\uparrow$ & SPL $\uparrow$ \\
\midrule
\method (FP16)  & 594.58 & 18.5 & & 5.37 & 57.6 & 49.7 & 45.5 \\
\rowcolor{myblue}
\method (W4A16)  & \bf367.80 & \bf8.6 & & 5.66 & 56.8 & 48.2 & 43.6 \\
\bottomrule
\end{tabular}}}
}
% \vspace{-0.5em}
\label{tab:quantization}
\end{table*}

\subsection{Limitations}
\label{sup:limitations}

\begin{figure*}[h]
  \centering
  \includegraphics[width=0.99\textwidth]{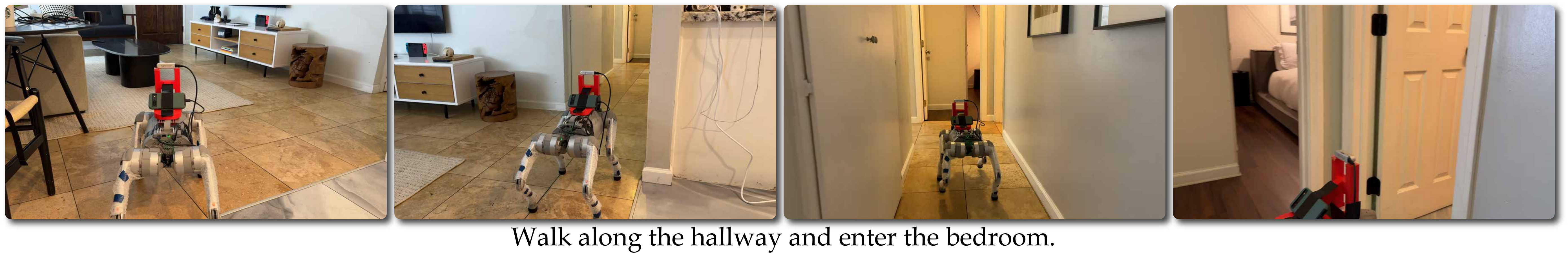}
  \caption{Failure case of \method.}
  \label{sup:failure}
\end{figure*}

In Figure~\ref{sup:failure}, we highlight a failure case in the real world where the robot initially follows the prompt but ultimately fails to reach the bedroom. This failure arises from the robot's inability to perform effective error correction when deviations occur. To further enhance performance, improving generalizability and spatial understanding is key. One potential direction is larger-scale training on more realistic simulations, which could provide more diverse navigation scenarios and error recovery cases. Additionally, incorporating explicit reasoning data during training could help the model better anticipate and correct mistakes.

